\documentclass{datologyai}
\usepackage{fontspec}

\usepackage{url}
\usepackage{makecell}   

\DeclareCaptionLabelSeparator{dlysep}{.\hspace{0.6em}}

\newcommand{\datnum}[1]{\textcolor[HTML]{002ECF}{\textbf{\boldmath#1}}}

\definecolor{datblue}{HTML}{002ECF}
\newcommand{\dcell}[1]{\textcolor{datblue}{#1}}
\newcommand{\famlogo}[1]{\,\raisebox{-0.25ex}{\includegraphics[height=1.5ex]{figures/logos/#1.png}}}

\newcommand{\cD}{\ensuremath{{\color[HTML]{2A61F1}D}}\xspace}
\newcommand{\cM}{\ensuremath{{\color[HTML]{8C959C}M}}\xspace}
\newcommand{\cQ}{\ensuremath{{\color[HTML]{7E22CE}Q}}\xspace}
\newcommand{\nDat}{\textcolor[HTML]{2A61F1}{Datology}}
\newcommand{\nMam}{\textcolor[HTML]{8C959C}{Mammoth}}
\newcommand{\nQwen}{\textcolor[HTML]{7E22CE}{Qwen3.5}}

\definecolor{baselinegray}{HTML}{2B2B2B}
\def\rolelabelstyle{2}
\newcommand{\curated}{\ifcase\rolelabelstyle curated\or \textit{curated}\or \textcolor[HTML]{002ECF}{\textit{curated}}\fi\xspace}
\newcommand{\Curated}{\ifcase\rolelabelstyle Curated\or \textit{Curated}\or \textcolor[HTML]{002ECF}{\textit{Curated}}\fi\xspace}
\newcommand{\base}{\ifcase\rolelabelstyle baseline\or \textit{baseline}\or \textcolor{baselinegray}{\textit{baseline}}\fi\xspace}
\newcommand{\Base}{\ifcase\rolelabelstyle Baseline\or \textit{Baseline}\or \textcolor{baselinegray}{\textit{Baseline}}\fi\xspace}

\newcommand{\papertitle}{Inducing Concision in VLMs via Data Curation}

\projectname{{\fontsize{22.5}{26}\selectfont\textcolor[HTML]{002ECF}{Brevity is the Soul of \\Inference Efficiency:}}}
\title{\papertitle}
\teambyline{DatologyAI Team\footnotemark}
\runningtitle{Brevity is the Soul of Inference Efficiency}

\abstract{%
Inference efficiency is typically pursued by shrinking the model: distillation, pruning, quantization, and sparse routing each lower per-token cost while treating token count as fixed. But output length has been inflating, and it is precisely the component the standard toolkit leaves untouched. Here, we argue that brevity is the missing inference-efficiency lever, and that pretraining data curation is a practical way to pull it: a model trained on concise, correct data learns to answer in fewer tokens; i.e. it has a lower \emph{Cost-of-Pass}. We apply our VLM curation pipeline to the MAmmoTH-VL single-image subset, and compare models trained on our curated data, the standard MAmmoTH-VL data, and external open-weight frontier VLMs. On a controlled 20-evaluation set and 14 VLMs at 1B--4B activated parameters, we hold output length fixed with a per-model regression, separating brevity from quality, and price models in FLOPs per correct answer. Curation buys a \datnum{$35\times$} Cost-of-Pass advantage over the most verbose 4B comparator (Qwen3.5-4B) within ${\sim}1$~pp of accuracy (0.41 vs 14.58 TFLOPs per correct answer; 0.691 vs 0.704 mean accuracy). Curation also buys a \datnum{$+17.55$-percentage-point} matched-length accuracy gain over the uncurated baseline, a gain that \datnum{grows with model scale} (from $+16.7$~pp at 1B to $+21.2$~pp at 4B). This brevity improvement concedes no quality: generic verbosity buys no accuracy at any capability or scale, and the window where reasoning-structured verbosity still earns its tokens shrinks from 4 of 8 capability groups at 2B to 1 of 8 at 4B. Per example, the concise model even reaches correct answers the verbose reasoning model misses, marking reasoning as a distinct curation target rather than something brevity gives up. Inference efficiency in this regime is a tokens-per-correct problem, and brevity is the lever that targets it directly.
}

\begin{document}

\maketitle
\footnotetext{See Contributions and Acknowledgements (\S~\ref{sec:contrib}) for full author list.}

\vspace{-5mm}
\begin{figure}[h]
  \centering
  \includegraphics[width=0.95\linewidth]{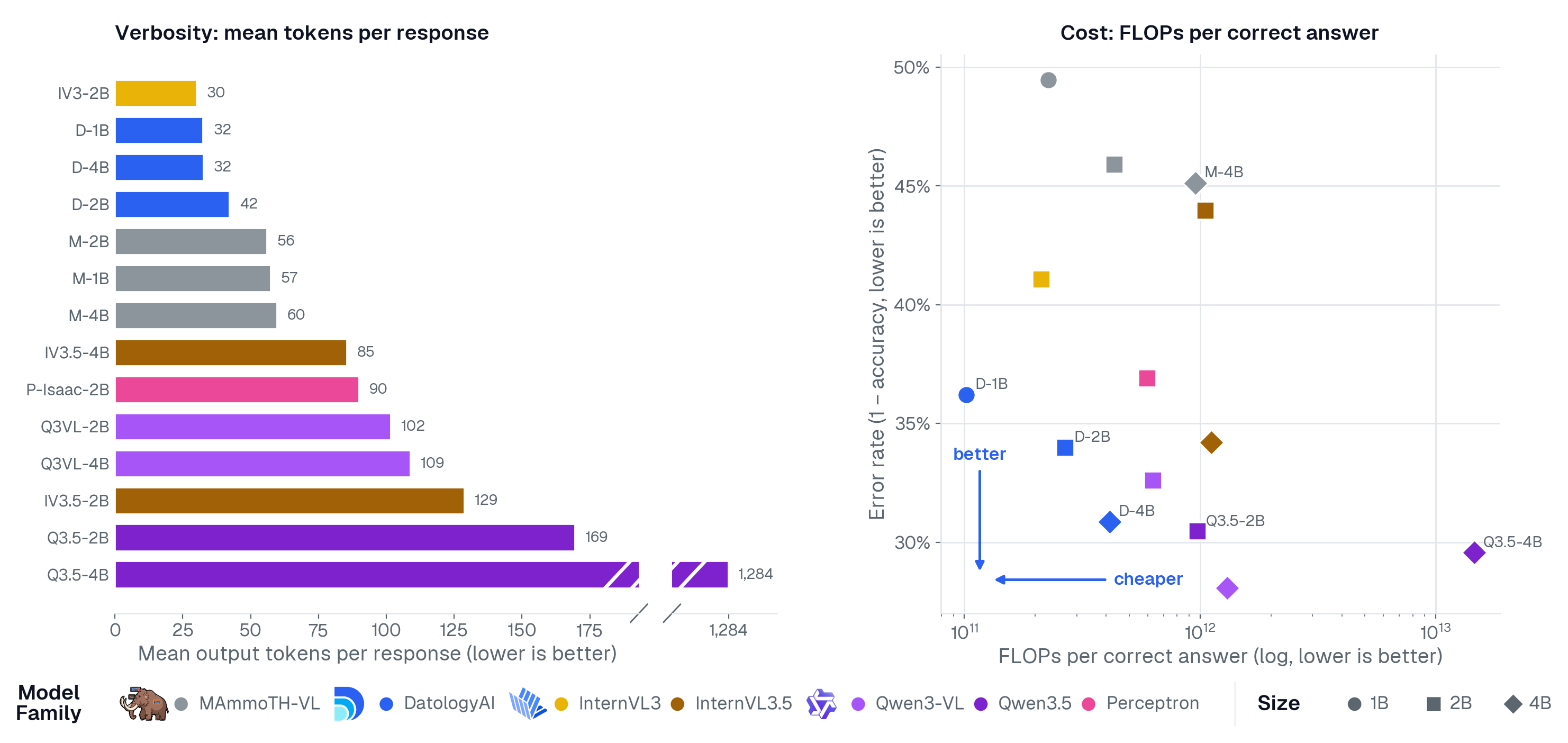}
  \caption{\textbf{The paper in one figure: brevity is the soul of inference efficiency, and data curation is how you enact it.} On a frontier pool of open-weight 2B--4B vision-language models, our \curated models (Datology, blue) answer in far fewer tokens than verbose comparators and cost far less per correct answer, with no loss of quality. \textbf{Left:} mean output length per response (the $x$-axis is broken to keep the cluster legible); the \curated models emit ${\sim}30$ tokens per response where Qwen3.5-4B emits $1{,}284$, a ${\sim}40\times$ gap. \textbf{Right:} error rate ($1{-}$accuracy) against FLOPs per correct answer; the \curated models hold the low-error, low-cost corner of the frontier and reach a \datnum{$35\times$} Cost-of-Pass advantage over Qwen3.5-4B within ${\sim}1$~pp of accuracy.}
  \label{fig:hero}
\end{figure}

\section{Introduction}
\label{sec:intro}

\begin{quote}
``I have made this letter longer than usual because I have not had the time to make it shorter.''\\
--- Blaise Pascal, \emph{Lettres provinciales} (1657)
\end{quote}

The conventional framing of inference efficiency is to \textbf{shrink the model}. The standard toolkit (distillation, pruning, quantization, MoE routing, speculative decoding, early-exit, and cascading) shares one implicit premise: the \emph{output} is a fixed quantity, and the only question is how cheaply each token of it can be produced. Optimize FLOPs per token; accept the token count. The metric we use throughout is \emph{Cost-of-Pass}: the compute, and ultimately the dollars, a model spends per correct answer (\S\ref{sec:cost}).

Shrinking the model is the usual move. A training-time gain can be cashed out along three fungible axes (speed, quality, and size) that together form the \emph{training-efficiency triangle}. A result sold as an inference-efficiency win has almost always been cashed out as size: fewer activated parameters. This leaves a fourth axis untouched: output length, which has remained an outcome when it could be an objective. It also sits off the standard scaling-law axes of compute, parameters, and training tokens, so a model can ship with a $10\times$ verbosity increase and no number on its model card will move, even as test-time compute becomes a scaling axis in its own right.

Pascal names what that premise misses. The long letter is the easy one to write; the short one takes work. Given the downward pressure on inference cost, the real question is not whether to pay for brevity but \emph{where}: at decode time, where length penalties and sampling controls are bolted onto a model not trained to be brief, or once and up front, in the data the model learns from.

The inference bill is coming due regardless, because token generation is increasing quickly, both in per-generation length and total quantity: \citet{epoch2025length} tracks roughly $2.2\times$ annual growth in mean output length on non-reasoning models and roughly $5\times$ on reasoning models, which now emit about $8\times$ more tokens than their non-reasoning peers. OpenAI ships reasoning-effort as an explicit product dial, where the default ``medium'' to ``high'' setting is a $1.6\times$ bump in output tokens~\citep{openaiReasoningEffort}. Sundar Pichai opened Google I/O 2026 with a token counter rather than a model: 9.7 trillion tokens per month in 2024, 480 trillion in 2025, \textbf{3.2 quadrillion today}, roughly $330\times$ growth in two years~\citep{pichai2026io}. The per-token gains from the standard toolkit are being recycled into longer outputs \emph{and} more queries, so total inference bills grow rather than shrink.\footnote{Thanks, Jevon}

A meaningful slice of reported quality gains is bought with test-time compute. The ``Price of Progress'' analysis \citep{priceofprogress2025} finds that roughly half of GPQA-Diamond progress at the frontier tracks increased inference spend, not algorithmic improvement; controlling for hardware, algorithmic efficiency improves at about $3\times$ per year, well below the headline 5--10$\times$. \citet{brown2026} presses the same case from inside a frontier lab: as models get better at spending test-time compute, a single benchmark number says less each release, and capability should be reported against a token or cost budget. \citet{singhal2024length} show that in chat RLHF a purely length-based reward reproduces most of the apparent gain of RLHF over SFT: length is not a side effect of RLHF, it is most of what RLHF does in the helpfulness setting. \citet{dubois2024length} show that controlling AlpacaEval for length shifts rankings substantially, with open-source RLHF'd models dropping the most. The common thread is that output length accounts for gains usually credited to better algorithms or training, not to length itself.

So there is a second axis worth optimizing: \textbf{how many tokens does the model spend to be correct?} At fixed model size and fixed accuracy, halving the output length roughly halves the dollar cost and the wall-clock latency, and more than halves the cost for long outputs, once attention's quadratic-in-length term is counted (\S\ref{subsec:m-cost}). No quantization scheme or speculative decoder can give this, because those reduce \emph{cost per token}; this reduces \emph{tokens per correct answer}. And the two reductions compose: parameter count and output length are independent factors of the inference bill, so a $2\times$ smaller model that is also $2\times$ less verbose is $4\times$ cheaper per correct answer, and a single curation move can drive both.

Verbosity can be attacked at several layers: output-layer length penalties in decoding \citep{wu2016google,murray2018correcting}, decoding controls and length-aware sampling \citep{holtzman2020curious,welleck2020neural}, RL-side length shaping \citep{singhal2024length,park2024disentangling}, and post-hoc summarization or two-pass generation (see \S\ref{sec:related} for the surveyed alternatives). This is typically where Pascal's letter gets shortened, and where the time gets spent.

We also argue for \textbf{inference efficiency through brevity, but achieved through data curation} at the pretraining stage. A data intervention internalizes the inference-time intervention into the model: trained on well-curated data, the model needs no external length constraint at decode time, and any such constraint then composes with that data-internalized brevity instead of fighting a model that was never trained to be brief.

This work is a follow-up to \citet{datology2026vlm}, which introduced our VLM data-curation pipeline applied to the public \textbf{MAmmoTH-VL} single-image training corpus \citep{guo2024mammothvl}. The models here are the same: DatologyAI models trained on the curated mixture, and models trained on the MAmmoTH-VL baseline, holding everything else fixed. We also evaluate against open-weights frontier models in the Qwen, InternVL, and Isaac families. The prior paper's headline was the \textbf{accuracy} effect of curation: $+11.7$~pp on average across a 20-eval suite at matched compute, with a side observation that response FLOPs were also lower at every scale. What it left unresolved is where the inference-cost win comes from: does it flow from \emph{fewer tokens at fixed accuracy} (brevity), \emph{more accurate tokens at fixed length} (quality at matched verbosity), or both? Does verbosity convey any advantages? Are there meaningful differences between reasoning and non-reasoning verbosity? The present work addresses these questions directly, using the same 20-eval VLM pool spanning 1B--4B parameter models:

\begin{itemize}
  \item \textbf{Curation cuts Cost-of-Pass.} Datology 4B answers correctly for \textbf{0.41 TFLOPs} against \textbf{14.58} for the most verbose 4B comparator, Qwen3.5-4B, within ${\sim}1$~pp of accuracy (0.691 vs 0.704): a \textbf{\boldmath$35\times$} gap. Per-token FLOPs are identical at fixed parameters by construction, so what separates the bills is tokens per correct answer (\S\ref{sec:cost}).
  \item \textbf{Accuracy holds at matched length.} At the same output length on the same benchmark, the curated model is correct \textbf{\boldmath$+17.55$~pp} more often than its uncurated paired baseline (pooled across scales), and the effect \textbf{grows with scale}, from $+16.7$~pp at 1B to $+21.2$~pp at 4B (\S\ref{sec:quality}).
  \item \textbf{Cheaper per correct, even when less accurate.} Against the verbose Qwen pool the curated model gives up 4--7~pp of length-controlled accuracy for the order-of-magnitude cost advantage; against both InternVL variants it wins on both axes. The Datology 4B vs Qwen3.5-4B pair can't be length-matched at all (2.2\% overlap in output length distributions): Qwen3.5-4B's accuracy is bundled with its chain-of-thought verbosity (\S\ref{sec:frontier}, \S\ref{sec:verbosity}).
  \item \textbf{Verbose tokens rarely earn their cost, and less so at scale.} Across concise non-reasoning (Datology), verbose non-reasoning (Mammoth), and verbose reasoning (Qwen3.5) models matched at 2B and 4B, \textbf{generic verbosity buys no accuracy anywhere}, while reasoning-structured verbosity beats curation on a shrinking subset of the eight capability groups: 4 of 8 at 2B, 1 of 8 at 4B, where the concise model instead wins 4 of 8 (\S\ref{sec:verbosity}).
  \item \textbf{Reasoning is a separate curation target from brevity.} Per example, the concise curated model reaches correct answers the verbose reasoning model misses, and vice versa. The two have different error sets, so curation does not simply compress reasoning into fewer tokens; reasoning is a separate capability to curate for (\S\ref{sec:verbosity}).
\end{itemize}

Together, these results demonstrate that \textbf{brevity is a viable inference-efficiency lever}, and that \textbf{data curation at pretraining is a viable, currently-underweighted access path to it}.

Pascal paid for brevity one letter at a time; curation pays once and collects on every generation thereafter.

\section{Related work}
\label{sec:related}

The verbosity confound has been studied only in pieces, scattered across text-only chat, reasoning models, VLMs, and, increasingly, the industry's macro cost picture. Five threads run through that literature, and our results plug into each: output length is climbing fast; longer responses are not reliably better ones, and in chat RLHF most of the apparent gain over SFT is length rather than content; reasoning models make the tradeoff concrete, coupling accuracy to a token bill the field is now learning to price; the same length confound almost certainly reaches VLMs, though it is barely measured there; and paying for brevity once, in the training data, is a form of amortized inference. Across all five, output length is an under-acknowledged confound in how model quality is read, and curation is a lever for it that almost no one is pulling.

\subsection{Text length is inflating, fast}
\label{subsec:related-length}

Model outputs are getting longer year over year, reasoning and non-reasoning models alike. \citet{epoch2025length} put numbers on the trend: across benchmark questions, non-reasoning models grow their mean output roughly $2.2\times$ per year and reasoning models roughly $5\times$. Their corpus is mostly text and reasoning models, but the inflation is broad enough that we take it as the backdrop for our vision-language results: cutting verbosity matters only because the field is busy inflating it.

\subsection{In chat RLHF, length explains most of the apparent gain}
\label{subsec:related-rlhf}

Longer answers can look better without being better. \citet{singhal2024length} give the cleanest precedent for this \emph{length-confound}: an apparent quality gain that, on inspection, is mostly a length gain rather than a content gain. In chat RLHF, a reward that depends on nothing but response length reproduces most of the downstream improvement of RLHF over SFT: the reward model inherits a length bias from the preference data it was fit on, so optimizing against it largely optimizes for length. \citet{dubois2024length}'s Length-Controlled AlpacaEval sharpens the point from the eval side: controlling for length raises agreement with the LMSYS Arena from 0.94 to 0.98, and a model's drop under that control becomes a measure of how much its score was riding length rather than quality, a measure which affects open RLHF'd models most. \citet{huang2024length} show the bias is both correctable and pervasive: length-residualization adds $+3.11$ to the RewardBench average across 33 reward models, and GPT-4-as-judge carries the same preference for verbosity, so it propagates wherever a judge is used. The shared claim, across all three, is that most of an apparent quality gap can turn out to be length. But the evidence is entirely chat-side and RLHF-shaped; whether the same artifact inflates quality numbers in vision-language models has not been tested.

\subsection{Reasoning models, cost-per-correct, and the longitudinal picture}
\label{subsec:related-reasoning}

A cluster of 2025--2026 work measures the unit economics of verbosity directly. OckBench \citep{ockbench2025}, which proposes \emph{per-token intelligence} as an evaluation axis, finds that same-size 7B reasoning models with similar accuracy can differ $3.3\times$ in tokens and $5\times$ in latency, and coins the ``Overthinking Tax'' for the verbosity of distilled small models. ``Decomposing Reasoning Efficiency'' \citep{decomposing2026} separates accuracy from token cost across 25 models on CogniLoad \citep{cogniload2025}, a synthetic reasoning benchmark with independently tunable chain-length, difficulty, and distractor density, and finds the two rankings only loosely aligned (Spearman $\rho = 0.63$): verbalization overhead, the share of tokens a model spends beyond what the answer needs, varies roughly $9\times$ across models and is only weakly tied to scale.

Nous Research's ``Measuring Thinking Efficiency in Reasoning Models'' (Aug 2025) gives the clearest industry-side statement of the problem \citep{nous2025thinking}: reasoning models spend hundreds of tokens on simple knowledge questions, and closed-weight frontier models are substantially more token-efficient than their open-weight peers.\footnote{That the most cost-exposed players, the closed-weight labs paying their own inference bills, are also the most token-efficient is itself a tell that brevity matters most to whoever hosts the model.} The ``Price of Progress'' analysis \citep{priceofprogress2025} is the headline cost number behind this post: frontier dollars-per-benchmark fall 5--10$\times$ per year, but algorithmic efficiency improves only ${\sim}3\times$, and roughly half of GPQA-Diamond progress at the frontier is bought with additional inference spend rather than better algorithms. \citet{erol2025costofpass}'s Cost-of-Pass supplies one of the units we adopt, the expected dollars per correct solution, and shows that inference-time techniques such as majority voting and self-refine rarely justify their cost.

\citet{brown2026} makes the case normatively: as models get better at spending test-time compute, benchmark performance keeps climbing without a clear plateau, so a single accuracy number is increasingly uninformative and models should be reported as performance-versus-compute curves rather than scalars. He also notes that raw token counts are not comparable across models (tokenizers, speeds, and per-token costs all differ), which is precisely why we denominate cost in FLOPs-per-correct (\S\ref{subsec:m-cost}).

Together these establish the units on which verbosity's price can be read directly: accuracy-per-token and cost-per-correct. What none of them measures is whether pretraining data curation moves those units.

\subsection{VLM-specific evidence, and the open gap}
\label{subsec:related-vlm}

The VLM-side evidence is thinner, but it points the same way. \citet{lee2025captioning} is the closest analog: as an MLLM's caption grows, the model leans progressively more on its own generated text and less on the image, and longer captions hallucinate more. They also show that strong VQA performance does not imply strong detailed-captioning performance, because VQA evals implicitly control for length while captioning does not, the same asymmetry our quality measurement (\S\ref{subsec:m-quality}) is built to respect. \citet{jung2025visual} show that hallucination-mitigation methods trade precision for recall on CHAIR (the standard object-hallucination metric for image captioning), so both are gameable by adjusting verbosity. And multimodal judges inherit the text-judge verbosity bias \citep{chen2024mllmjudge}, so any VLM eval routed through a judge carries the confound. Unfortunately there is no clean VLM analog to Singhal et al.'s result that a purely length-based reward reproduces most of the RLHF gain. We position \S\ref{sec:quality}'s length-controlled accuracy regression and hallucination decomposition to be that analog from the data-curation side.

\subsection{Amortized inference: brevity as cached cognition}
\label{subsec:amortized}

Cognitive science already has a name for paying a cost once so you needn't re-derive an expensive answer every time: \emph{amortized inference} \citep{gershman2014amortized}. The basic principle is that an intelligent system avoids re-deriving expensive inferences from scratch and instead learns a cheap, reusable approximation. Through that lens, a short correct answer from a well-trained model is amortized inference: the expensive deliberation happened during training, and the forward pass is the cached approximation.\footnote{Bayesian amortized inference has a precise technical meaning (training a network to approximate an intractable posterior), and we are not doing exactly that; the analogy is to the direction of the work, not the formalism.} Data curation is the learning signal that decides what gets amortized: show the model concision and it learns to produce the answer directly rather than re-deriving it token by token at decode time. The everyday version of the same move is the shift from System 2 to System 1: a deliberate, multi-step computation, run often enough, compiles into a fast automatic response, the way a chess expert sees a strong move at a glance instead of searching for it.

This reframes the cost question. Test-time compute is un-amortized inference: the model paying for deliberation at every query, every user, every time. Curation amortizes that cost into the weights, paid once at training and spread across every generation the model later produces. It also turns inference cost into a training signal: where production usage shows the model repeatedly spending tokens on a recurring pattern, that pattern becomes a curation target for the next pass, and the next model is leaner still, a data flywheel in which deployment cost feeds back into the training set. And it meets the obvious Bitter Lesson objection \citep{sutton2019bitter} in its own terms: curation does not add hand-built deliberation, it amortizes the human-rater and prior-model compute already latent in the data, which is exactly the kind of cost a scalable method is meant to absorb.

The analogy flatters deep learning research in one respect. Biological amortization is mostly undirected: evolution compiles useful priors into reflexes over many generations, and within a lifetime the brain caches whatever the world happens to present, the way a hidden figure in a two-tone image, once resolved, is seen instantly and permanently from a single glance. Deep learning gets to amortize deliberately: we choose which inferences get compiled into the weights, across the whole training distribution, on every run. We are not waiting for evolution, or for the right experience to come along; we are running amortization on demand.

\section{Results}\label{sec:results}

\subsection{Cost-of-Pass: thoughtful data curation is Pareto-dominant on inference cost}
\label{sec:cost}

Inference efficiency is conventionally reported in dollars per million tokens. That prices the tokens a model emits but says nothing about how many it needs, or whether they are right. Two quantities actually drive the bill: \textbf{cost per answer} (the compute each response burns) and \textbf{accuracy} (how often that response is correct), and we report both throughout. We also summarize them with a single ratio, \textbf{cost per correct answer} (cost per answer divided by accuracy), which has the right units for what a correct answer costs. We treat it as a convenient aggregate, not the one true metric: dividing by accuracy prices a point of accuracy linearly, and whether that exchange rate matches an application's value of being right is a judgment we leave to the reader.

We measure cost per answer in FLOPs, via the decode-dominated proxy of \S\ref{subsec:m-cost}: a response of $\bar{n}_{\text{out}}$ tokens from a dense model with $N$ active parameters costs about $2N\bar{n}_{\text{out}}$ FLOPs, so cost per correct $= 2N\bar{n}_{\text{out}}/\text{accuracy}$. This proxy is faithful to the small dense VLMs we study, where decode compute scales cleanly with active parameters and output length, and a rough lower bound elsewhere: it omits attention, which is quadratic in sequence length, along with the bandwidth and memory limits that bite at longer sequences and larger or sparsely-activated models (\S\ref{subsec:m-cost}). Because the omitted attention cost grows with output length, the proxy understates the verbose models' cost the most, so the Cost-of-Pass gaps we report are conservative. This quantity is hardware-independent; a dollar translation under a self-hosted H100 roofline is in \S\ref{subsec:m-cost}.

\paragraph{Headline arithmetic: the \curated model is cheapest per correct at every matched scale.}
Table~\ref{tab:flops-per-correct} reports FLOPs per correct across 14 open-weight VLMs at 1B--4B parameters, averaged over the 20-eval frontier suite (\S\ref{subsec:m-pool}). Rows are ordered by FLOPs/correct ascending.
\begin{table}[h]
  \centering
  \small
  \begin{tabular}{l rrrr}
    \toprule
    Model & $N$ (B) & $\bar{n}_{\text{out}}$ & Accuracy & FLOPs/correct (TFLOPs) \\
    \midrule
    \dcell{Datology 1B}\famlogo{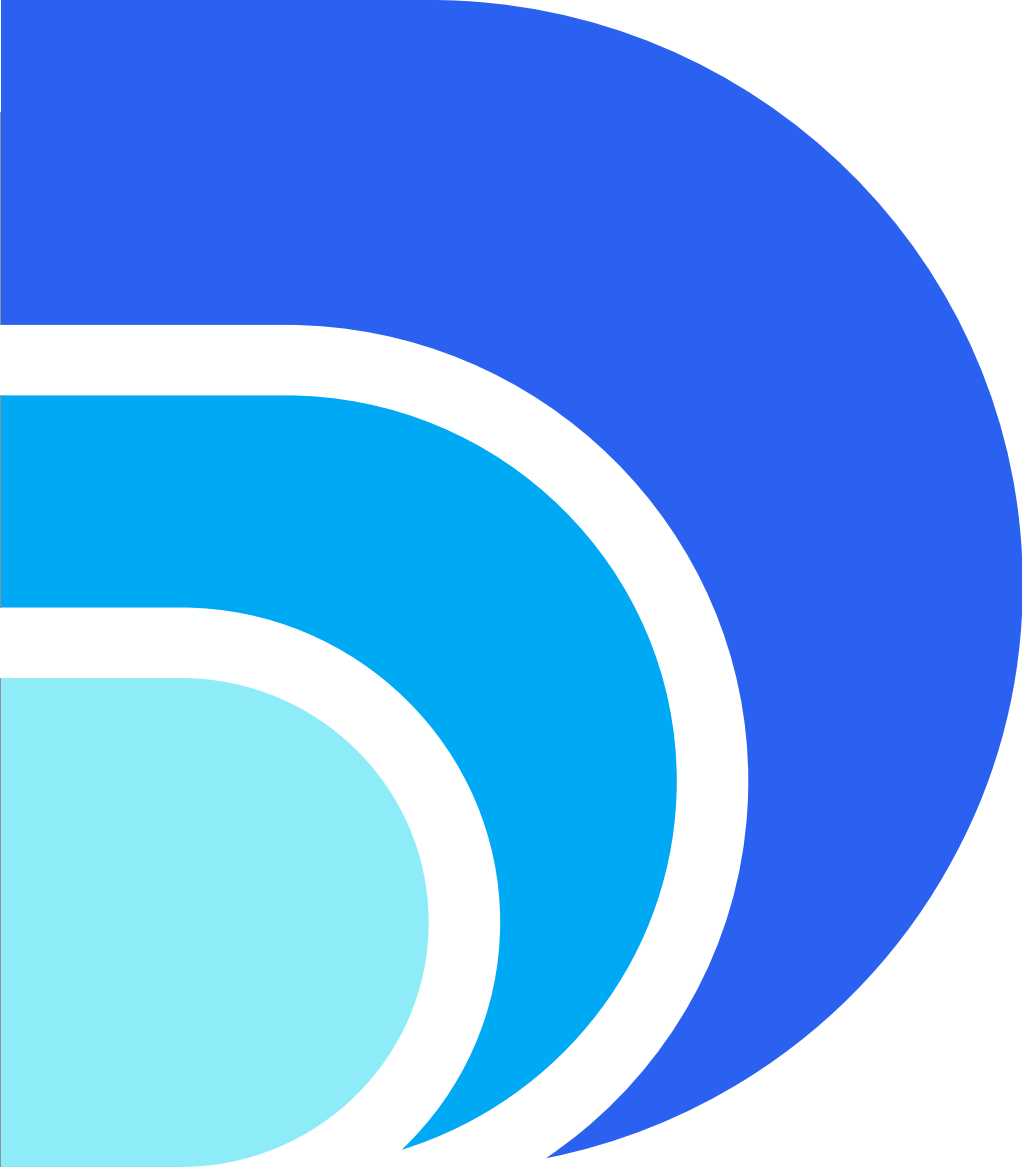}          & \dcell{1.0} & \dcell{32.4}  & \dcell{0.638} & \dcell{0.10} \\
    InternVL3-2B\famlogo{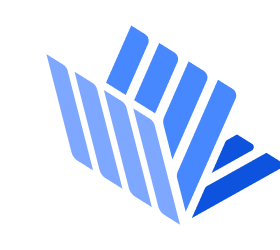}        & 2.1 & 30.0  & 0.589 & 0.21 \\
    Mammoth 1B\famlogo{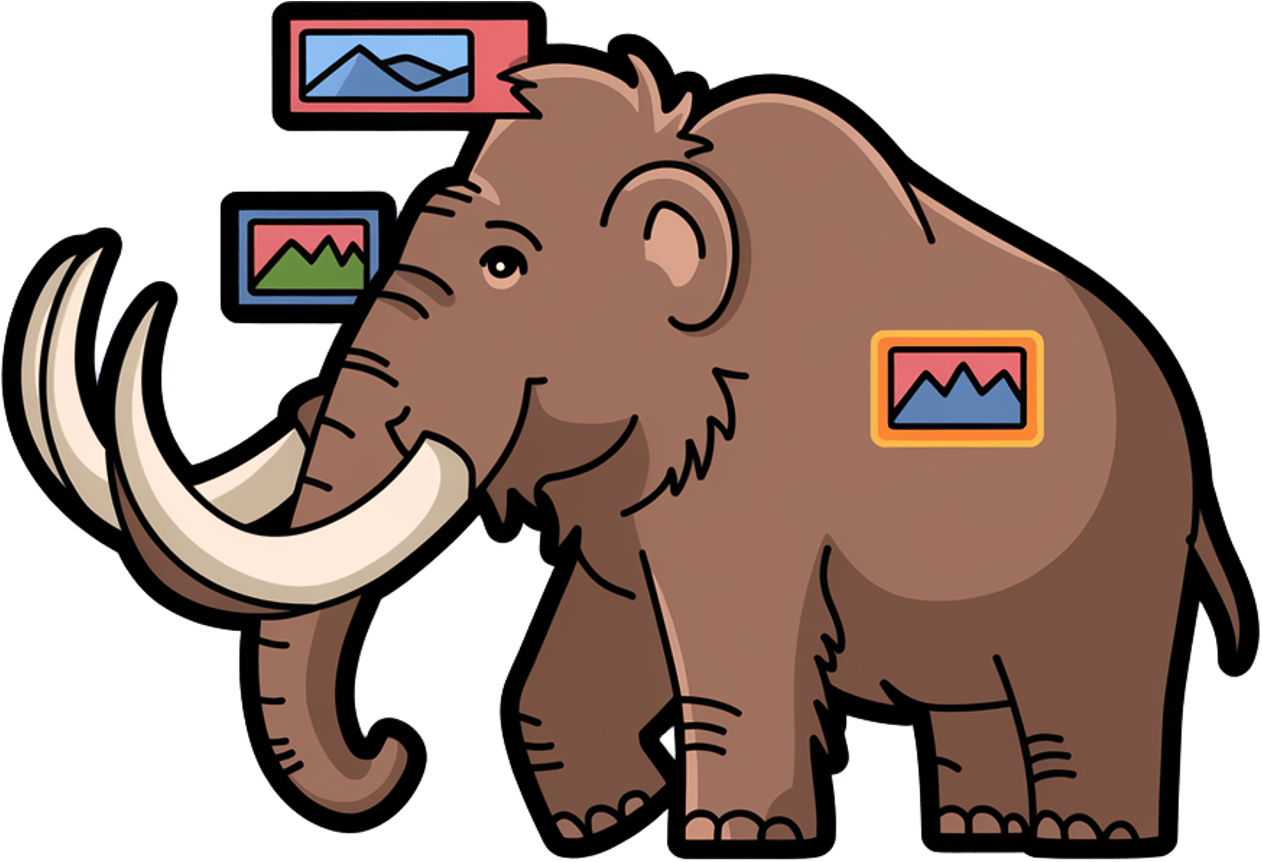}           & 1.0 & 57.3  & 0.505 & 0.23 \\
    \dcell{Datology 2B}\famlogo{datology}          & \dcell{2.1} & \dcell{42.1}  & \dcell{0.660} & \dcell{0.27} \\
    \dcell{Datology 4B}\famlogo{datology}          & \dcell{4.4} & \dcell{32.5}  & \dcell{0.691} & \dcell{0.41} \\
    Mammoth 2B\famlogo{mammoth}           & 2.1 & 55.8  & 0.541 & 0.43 \\
    Perceptron-Isaac-2B                   & 2.1 & 89.8  & 0.631 & 0.60 \\
    Qwen3-VL-2B-Instruct\famlogo{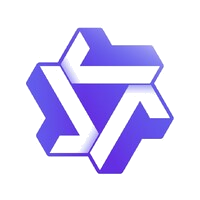} & 2.1 & 101.6 & 0.674 & 0.63 \\
    Mammoth 4B\famlogo{mammoth}           & 4.4 & 59.6  & 0.549 & 0.96 \\
    Qwen3.5-2B\famlogo{qwen}           & 2.0 & 169.4 & 0.695 & 0.97 \\
    InternVL3.5-2B\famlogo{internvl}      & 2.3 & 128.6 & 0.560 & 1.06 \\
    InternVL3.5-4B\famlogo{internvl}       & 4.3 & 85.4  & 0.658 & 1.12 \\
    Qwen3-VL-4B-Instruct\famlogo{qwen} & 4.3 & 108.9 & 0.719 & 1.30 \\
    Qwen3.5-4B\famlogo{qwen}           & 4.0 & 1{,}284 & 0.704 & 14.58 \\
    \bottomrule
  \end{tabular}
  \caption{\textbf{The \curated models cluster at the cheapest end of the Cost-of-Pass frontier.} Datology and Mammoth rows are paired by scale (matched controls); at every scale the \curated model is both cheaper per correct and more accurate than its control, and the \curated checkpoints are three of the five cheapest rows. The one model that undercuts a \curated checkpoint on cost, InternVL3-2B (0.21 vs Datology 2B's 0.27~TFLOPs), does so only by giving up ${\sim}10$~pp of accuracy (\S\ref{sec:frontier}). FLOPs computed as $2N\bar{n}_{\text{out}}/\text{accuracy}$ per \S\ref{subsec:m-cost}.}
  \label{tab:flops-per-correct}
\end{table}

Both curation and verbosity are demonstrated paths to better models: curation produces Pareto improvements (cheaper \emph{and} more accurate); verbosity produces accuracy gains at the cost of more tokens. The matched-scale pairs measure the magnitude of the first; the frontier comparators show what the second costs.

At every paired scale, the \curated model is \emph{both} cheaper per correct and more accurate than its Mammoth control. At 1B, the \curated model costs $2.2\times$ fewer FLOPs per correct (0.10 vs 0.23 TFLOPs) and is 13.3~pp more accurate. At 2B, $1.6\times$ fewer FLOPs (0.27 vs 0.43) and 11.9~pp more accurate. At 4B, $2.3\times$ fewer FLOPs (0.41 vs 0.96) and 14.2~pp more accurate.

Against the frontier-comparable external models, the \curated 4B at 0.41 TFLOPs/correct sits below every comparator at its weight class: 2.7--3.2$\times$ below InternVL3.5-4B and Qwen3-VL-4B (both within 3~pp of its accuracy), and $35\times$ below Qwen3.5-4B within ${\sim}1$~pp of accuracy (0.691 vs 0.704). Notably, Datology 4B is cheaper per correct than Mammoth 2B (0.41 vs 0.43 TFLOPs) despite carrying more than twice the activated parameters --- a direct demonstration that the brevity lever can overpower the capacity lever at this scale.

\paragraph{Verbosity is the bill.}
The cost separation at fixed scale cannot be explained by per-token efficiency. At 4B activated parameters, the FLOPs per output token ($= 2N$) are (near) identical across Datology, Mammoth, Qwen3.5, Qwen3-VL, and InternVL3.5. The variable that differs is the number of tokens emitted to reach a correct answer: 32.5 for Datology, 59.6 for Mammoth, 85.4 for InternVL3.5, 108.9 for Qwen3-VL, and 1{,}284 for Qwen3.5-4B.\footnote{Qwen3.5-4B is a reasoning-style model whose default inference setting emits long chain-of-thought on the math, chart, and counting benchmarks in the suite. Disabling reasoning mode or capping output tokens would lower this mean at a likely non-trivial accuracy cost on the reasoning-heavy subset; the value reported here is the model's default-configuration behavior, which is what the inference bill reflects on a deployed pipeline.} The $35\times$ gap in FLOPs/correct between Datology 4B and Qwen3.5-4B is roughly the $40\times$ ratio of their mean output lengths, attenuated by Qwen3.5-4B's slightly higher accuracy. Verbosity, not capacity, is the cost variable at this scale.

\paragraph{\boldmath Compounding: a \curated 1B is 143$\times$ cheaper than Qwen3.5-4B.}
The brevity lever and the capacity lever stack multiplicatively when both are pulled. Datology 1B at 1.0~B activated parameters and 32.4 mean output tokens lands at 0.10 TFLOPs/correct, the lowest in the table, $9.4\times$ below Mammoth 4B and \textbf{\boldmath$143\times$} below Qwen3.5-4B. The trade-off is 5--7~pp of accuracy below the larger Datology checkpoints (0.638 vs 0.660 and 0.691 for 1B, 2B, and 4B, respectively), which a deployment may or may not be willing to absorb depending on the application. The compounding is what makes the curated 1B competitive: a small model trained on thoughtfully-curated data sits, in inference-cost terms, an order of magnitude below frontier 4B models.

FLOPs per correct already prices each model with its parameter count folded in. Two further views confirm the gap isn't an artifact of that accounting: tokens per correct answer strips the parameters back out, isolating the share of the gap that is pure verbosity, and a dollar translation puts it in operating-cost terms.

\paragraph{\boldmath Tokens per correct, and dollars: the same ${\sim}40\times$ gap.}
OckScore (OckBench's per-token-intelligence framing, \S\ref{subsec:related-reasoning}; \citealp{ockbench2025}) places the same comparison on a hardware-free axis: tokens per correct answer.

\begin{figure}[h]
  \centering
  \includegraphics[width=\linewidth]{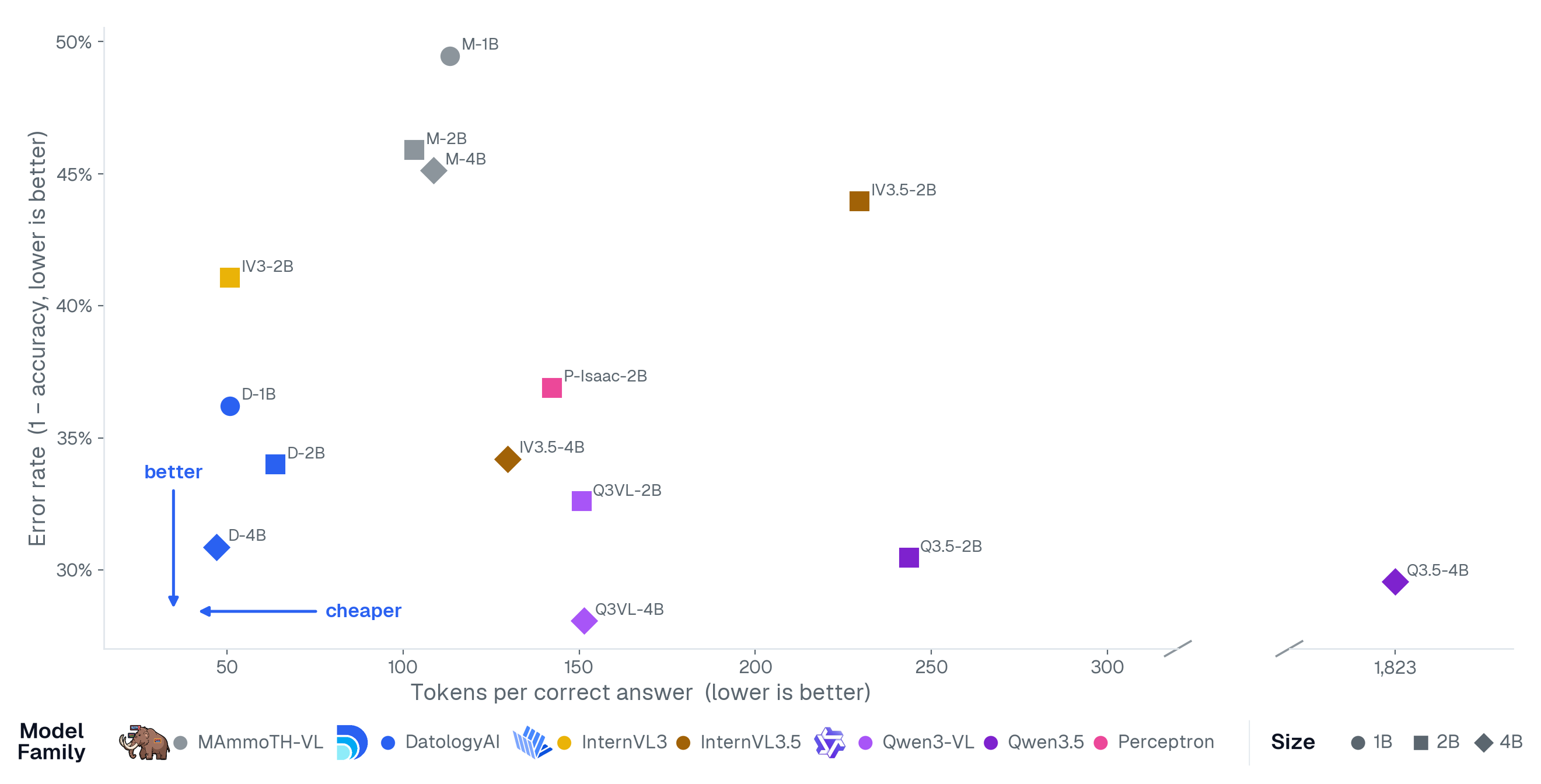}
  \caption{\textbf{OckScore: the \curated models own the low-cost, low-error corner.} Tokens per correct answer ($x$, broken to keep the cluster legible) against error rate ($y$); both axes run so that lower is better, placing the best models at the bottom-left. The \curated checkpoints cluster there at ${\sim}47$--$64$ tokens per correct, while the verbose comparators fan up and to the right, with Qwen3.5-4B alone at ${\sim}1{,}820$, a ${\sim}30$--$40\times$ token gap at comparable error.}
  \label{fig:ockscore}
\end{figure}

Datology 4B at ${\sim}47$ tokens/correct compares to ${\sim}244$ for Qwen3.5-2B and ${\sim}1{,}823$ for Qwen3.5-4B (full per-model table in \Cref{tab:tokens-per-correct}), the same ${\sim}40\times$ ratio the matched-scale view reports above. Because the dollar version of Cost-of-Pass is a constant-$N$ affine transform of token counts, the ratio must transfer; what OckScore confirms is that the cost gap is not a hardware-pricing artifact.

Translated to dollars under the \S\ref{subsec:m-cost} H100 roofline, the same ratios hold in operating-cost terms: a million correct answers cost about \$1.34 from the \curated 4B and \$47 from Qwen3.5-4B (and \$0.33 from the \curated 1B), roughly the price of a soda versus a tank of gas. Full per-model figures and the roofline arithmetic are in \S\ref{subsec:m-cost}; the dollar cost is linear in FLOPs/correct at fixed batch and bandwidth, so the dollar ratios equal the FLOPs ratios and the $35\times$ gap is invariant to reasonable hardware assumptions.

\begin{takeaway}
\textbf{Thoughtful data curation is Pareto-dominant on inference cost.} At every matched scale, the \curated model raises accuracy while lowering FLOPs-per-correct versus its uncurated control ($1.6\times$--$2.3\times$ FLOPs reduction; 11.9--14.2~pp accuracy gain). At 4B, the \curated model sits below every external comparator at its weight class --- 2.7$\times$--3.2$\times$ below InternVL3.5-4B and Qwen3-VL-4B at comparable accuracy, and $35\times$ below Qwen3.5-4B within ${\sim}1$~pp of accuracy.
\end{takeaway}

\subsection{Curation raises matched-length accuracy, more so at scale}
\label{sec:quality}

The previous section's cost claim rests on an assumption it hasn't yet tested: that accuracy on the eval suite reflects accuracy at any output length. If curation's accuracy gain over its uncurated baseline is really just shorter answers being easier to score, that gain (and the cost advantage built on it) is a metric artifact. To resolve this confound, we fit a length-controlled regression on the pairs of models trained on curated vs uncurated data. A separate worry, that brevity flatters the curated model against external models, is taken up in \S\ref{sec:frontier}: at matched length the verbose Qwen models are \emph{more} accurate than the \curated model, not less, so its cost-per-correct edge isn't bought with easier-to-grade short answers.

The setup holds every confound except curation fixed. Datology and Mammoth are paired at three scales (1B, 2B, and 4B). Within each pair: same base LLM, same VL training recipe, same context length, same eval pipeline. Only the pretraining data curation differs.

\paragraph{Curated models are correct 17.6 pp more often after controlling for length.}
We fit a logistic GLM on per-model output tokens, benchmark fixed effects, and scale fixed effects, pooled across all three matched scales:
\begin{equation*}
  \operatorname{logit} \Pr(\text{correct}) = b_0 + c \cdot \text{is\_datology} + \operatorname{cr}(\text{length\_tokens}, \mathrm{df}=4) + C(\text{benchmark}) + C(\text{scale})
\end{equation*}
on every per-example generation across the 18 grouped frontier evals (the 20-eval suite minus the two aggregate-only evals, BLINK and BrandID; \S\ref{subsec:m-pool}), with standard errors clustered at the example level. The coefficient $c$ is converted to an \textbf{average marginal effect} (AME) \textbf{in percentage points}: the expected change in the probability of a correct answer from switching the curated indicator on, averaged over the observed data, so the headline reads as a direct pp gap (\Cref{tab:ame-pooled}). A detailed methodological description is in appendix (\S\ref{subsec:m-quality}).

\begin{table}[h]
  \centering
  \small
  \begin{tabular}{l r}
    \toprule
    Quantity & Value \\
    \midrule
    \textbf{Length-controlled accuracy gap (AME, pooled across scales)} & \textbf{\boldmath$+17.55$~pp} \\
    95\% CI                                          & $[+13.33, +21.77]$ \\
    $z$-statistic                                    & 8.1 \\
    $N$ rows / $N$ clusters (\texttt{example\_id})   & 1{,}019{,}035 / 49{,}828 \\
    LPM cross-check (HC1)                            & $+17.79$~pp \\
    Sensitivity: same regression in chars (LPM HC1)  & $+17.91$~pp \\
    \bottomrule
  \end{tabular}
  \caption{\textbf{Length-controlled accuracy gap for the curated indicator}, pooled across the three matched scales. The linear-probability-model (LPM) cross-check refits the same specification with ordinary least squares and HC1 robust errors; the sensitivity row reruns the regression with output length measured in characters rather than tokens. Both agree with the logistic AME to within $0.4$~pp. The GLM specification is in \S\ref{subsec:m-quality}.}
  \label{tab:ame-pooled}
\end{table}

\paragraph{Reading.} At the same output length, on the same benchmark, the \curated model is correct ${\sim}17.6$ percentage points more often than its uncurated paired baseline, averaged across the three matched model sizes. Curation induces both brevity \emph{and} accuracy: the \curated models are short kings. The length-accuracy spline, the GLM's smooth fit of accuracy against output length (\Cref{fig:length-spline-controlled}), shows why the controlled and raw gaps nearly coincide: both models' outputs cluster at the short, flat end of the curve, so length contributes almost nothing to the raw gap.

\paragraph{Curation imparts greater benefit for larger models.}
Per-model-size regressions (one fit per matched scale, no scale fixed effect) reveal real heterogeneity: the curation effect is meaningfully larger at 4B ($+21.2$~pp) than at 2B ($+15.6$~pp), a ${\sim}6$~pp gap that is tightly estimated at both scales, with 1B in between ($+16.7$~pp). This tracks \S\ref{sec:cost}'s matched-scale FLOPs-per-correct ratios, which are also largest at 4B ($2.3\times$) and smallest at 2B ($1.6\times$). Two views of the same scaling story.

\paragraph{Cost arithmetic and length-controlled accuracy track each other.}
\S\ref{sec:cost}'s matched-scale cost arithmetic (\$0.87 vs \$1.40 per million correct answers at 2B, a $1.6\times$ ratio) is the same claim seen through the dollar denominator. The Cost-of-Pass numerator is correct answers per generation, so the $+17.6$~pp pooled AME is what makes the cost gap exist. Without the length-controlled accuracy edge, the matched-scale cost comparison would dissolve to ``same accuracy at same length, same cost.'' Instead, the \curated model produces between 16 and 21 percentage points more correct answers per generation at every scale, at fewer tokens per generation, on the same compute budget.

The per-scale FLOPs/correct ratios (Mammoth $\div$ Datology, from \S\ref{sec:cost}) and the per-scale length-controlled AMEs track each other across scales (\Cref{tab:cost-quality-track}).

\begin{table}[h]
  \centering
  \small
  \begin{tabular}{l rrrr}
    \toprule
    Scale & Mammoth FLOPs/correct & Datology FLOPs/correct & FLOPs cost ratio & Length-controlled AME (pp) \\
    \midrule
    1B & 0.23 TFLOPs & 0.10 TFLOPs & $2.2\times$ & $+16.73$ \\
    2B & 0.43 TFLOPs & 0.27 TFLOPs & $1.6\times$ & $+15.57$ \\
    4B & 0.96 TFLOPs & 0.41 TFLOPs & $2.3\times$ & $+21.20$ \\
    \bottomrule
  \end{tabular}
  \caption{\textbf{Cost and quality track each other across scales.} Per-scale FLOPs-per-correct ratios (from \S\ref{sec:cost}) alongside the per-scale length-controlled AMEs. Both are largest at 4B and smallest at 2B.}
  \label{tab:cost-quality-track}
\end{table}

Curation pays off more, in both quality and inference cost, at the larger scales, across two independent measurement axes.

\paragraph{Captioning quality holds, and the hallucination win is brevity.}
The pooled AME folds in DetailCaps via an LLM-judge precision signal (\S\ref{subsec:m-judge}); long-form captioning is the subset where correctness is hardest to pin down (captions are scored by judged precision and recall rather than exact match), so it warrants its own audit. On the controlled 2B pair, curation improves both DetailCaps precision ($+0.018$) and CAPability recall ($+0.006$) while cutting caption length by 59\% and hallucinated mentions per caption by 19\%. A length-versus-rate decomposition attributes the hallucination win to brevity rather than to grounding each claim more faithfully: per character, the \curated model's grounding is similar to Mammoth's. The full audit, table, and decomposition are in Appendix~\ref{app:captioning}.

\begin{takeaway}
\textbf{At the same output length, on the same benchmark, the \curated model is correct \boldmath${\sim}17.6$~pp more often than its uncurated paired baseline.} The $35\times$ cost advantage isn't a length artifact --- the length-controlled accuracy edge holds at every matched scale we tested, \emph{growing} from $+16.7$~pp at 1B to $+21.2$~pp at 4B (95\% CI on the pooled effect: $[+13.3, +21.8]$). Curation buys quality and efficiency at every matched scale, and its impact grows with model size.
\end{takeaway}

\subsection{Cheaper by the answer: cost-efficient against six of seven frontier-comparable models}
\label{sec:frontier}

The previous analyses showed that our curation substantially improves length-controlled accuracy relative to the baseline Mammoth data. But does the curated model's accuracy edge survive against external frontier models trained under different recipes, with different verbosity priors? Here we widen the lens to the frontier-comparable pool of open-weight 2B and 4B external models, where everything but parameter count varies. Across that pool the curated model holds the cost-efficient Pareto front against six of seven comparators, giving up ground at fixed length only to models that cost several times more per correct answer.

\paragraph{Length-controlled accuracy: competitive at fixed length, decisive on cost.}
We re-run the same length-controlled regression from \S\ref{sec:quality}, now pairwise against each external model, restricting to pairs whose output-length distributions overlap meaningfully (at least 10\% of one model's responses fall within the other's $[p10, p90]$ range). The Datology 4B vs Qwen3.5-4B pair fails that test at 2.2\% overlap; the next subsection shows why that failure is itself a result.

The average marginal effect (AME) is best interpreted as a controlled counterfactual, not a deployment number. A $-7$~pp AME against Qwen3.5-2B means that \emph{if both models were constrained to emit the same number of tokens}, Qwen would be roughly 7~pp more accurate on the overlapping range. But the models are not deployed at equal length, and forcing equal length would erase the brevity that makes the curated model cheap in the first place. At their actual operating points the curated model produces more correct answers per FLOP (\S\ref{sec:cost}): the AME is the stress test, Cost-of-Pass is the bill. On that stress test the curated model wins outright against four of seven external models (both InternVL3.5 variants, InternVL3, and Perceptron) and trails three --- the Qwen family --- by 4--7~pp, every one of which it still beats decisively on cost, as \Cref{tab:frontier-cost} sets out. \Cref{fig:frontier-spline} plots the per-pair length-accuracy splines behind these AMEs.

\paragraph{Qwen3.5-4B has no concise operating point.}
A length-controlled comparison needs both models to actually appear at the same output lengths; for Datology 4B and Qwen3.5-4B, they almost never do. The curated model's median output is 3 tokens against Qwen3.5-4B's 626, and only 2.2\% of Qwen3.5-4B's responses fall inside the curated model's central $[p10, p90]$ length range (\Cref{tab:positivity}).

Per benchmark, the disparity is even sharper. On 3DSRBench, Qwen3.5-4B outputs are \textbf{\boldmath$1{,}276\times$ longer} than Datology's; on AI2D, $1{,}222\times$ longer; on MathVista, $868\times$ longer; on MMBench, $716\times$ longer. The two models don't operate at comparable lengths in our data.

\textbf{This is the verbosity claim made concrete.} Qwen3.5-4B's accuracy comes bundled with its chain-of-thought verbosity. There is no ``concise Qwen3.5-4B'' operating point in the data: to access Qwen3.5-4B's quality, you have to pay the ${\sim}1{,}200$-token-per-response inference bill. The \S\ref{sec:cost} cost number ($35\times$ more FLOPs per correct answer than the curated 4B within ${\sim}1$~pp of raw accuracy) is the direct consequence of that bundling. Asking ``but what if you forced Qwen3.5-4B to be shorter?'' is asking a counterfactual the data are silent on, because Qwen3.5-4B is \emph{never observed shorter} in the operating mode it's deployed in.

\paragraph{Cost-of-Pass: cheaper per correct against six of seven.}
The accuracy gap is one half of the trade; cost is the other. \Cref{tab:frontier-cost} pairs the two at the same self-hosted H100 roofline from \S\ref{sec:cost} (\$2.5/hr, batch 64).

Recall that the AME is the accuracy gap at \emph{matched} output length: positive means the \curated model is more accurate at the same length, negative means the comparator is.
\begin{table}[h]
  \centering
  \small
  \setlength{\tabcolsep}{4pt}
  \begin{tabular}{l r r r r l}
    \toprule
    Comparator & AME (pp) & \makecell{Length\\overlap} & \makecell{Datology\\FLOPs/correct} & \makecell{Comparator\\FLOPs/correct\footnotemark} & \makecell{Cost\\ratio} \\
    \midrule
    vs Qwen3.5-2B            & $-6.99$  & 48\% & 0.27 TFLOPs & 0.97 TFLOPs & $3.6\times$ more expensive \\
    vs Qwen3-VL-2B-Instruct  & $-3.93$  & 89\% & 0.27 TFLOPs & 0.63 TFLOPs & $2.4\times$ more expensive \\
    vs Perceptron-Isaac-2B   & $+2.39$  & 18\% & 0.27 TFLOPs & 0.91 TFLOPs & $2.2\times$ more expensive \\
    vs InternVL3.5-2B        & $+9.68$  & 26\% & 0.27 TFLOPs & 1.05 TFLOPs & $3.9\times$ more expensive \\
    vs InternVL3-2B          & $+10.23$ & 86\% & 0.27 TFLOPs & 0.21 TFLOPs & $0.8\times$ --- \emph{cheaper} \\
    vs Qwen3-VL-4B-Instruct  & $-4.53$  & 81\% & 0.41 TFLOPs & 1.30 TFLOPs & $3.2\times$ more expensive \\
    vs InternVL3.5-4B        & $+5.69$  & 53\% & 0.41 TFLOPs & 1.12 TFLOPs & $2.7\times$ more expensive \\
    \bottomrule
  \end{tabular}
  \caption{\textbf{Length-controlled accuracy and Cost-of-Pass against the frontier-comparable pool.} Per comparator: the length-controlled AME (positive favors the \curated model), the length-distribution overlap, and self-hosted FLOPs-per-correct at the \S\ref{sec:cost} H100 roofline (cost ratio is comparator $/$ Datology). The \curated model is cheaper per correct than six of seven; only InternVL3-2B (smaller and shorter) is cheaper, at 10~pp lower fixed-length accuracy. The Perceptron-Isaac-2B row carries a low-overlap caveat (18\%); 95\% CIs for the AMEs are in \Cref{tab:frontier-ci}.}
  \label{tab:frontier-cost}
\end{table}
\footnotetext{Computed from per-model 20-eval macro mean tokens and the same $2N$ FLOPs-per-token roofline as \S\ref{sec:cost}.}

Against the Qwen family the curated model gives up 4--7~pp of length-controlled accuracy but costs 2.4--3.6$\times$ less per correct answer. Against the InternVL3.5 family --- and, narrowly, Perceptron --- it wins on \emph{both} axes at once, 2--10~pp more accurate \emph{and} 2.2--3.9$\times$ cheaper. The single inversion is InternVL3-2B: a smaller, shorter model that is 25\% cheaper per correct but 10~pp less accurate at fixed length, the one comparator that buys its cost edge with accuracy the curated model declines to give up. (The \S\ref{sec:cost} operating-point comparison against Qwen3.5-4B, excluded here for non-overlap, is the most lopsided of all: $35\times$ cheaper per correct within ${\sim}1$~pp of accuracy. The unconstrained per-example view of that pair is \S\ref{sec:verbosity}.)

One confound could undercut all of this: if the \curated model's shorter outputs came from refusing to answer more often, its length advantage would be a behavioral shift, not genuine brevity. It is not. A three-term decomposition of the mean-length gaps above attributes 98--100\% of each to \emph{response-conditional} brevity (shorter answers when the model does answer); the refusal-rate shift contributes at most 1.8\%, often negative, and the \curated model essentially never refuses on standard evals. The result survives a refusal-enriched stress test (Appendix~\ref{app:refusal}).

\begin{takeaway}
\textbf{Against the frontier-comparable pool, brevity buys cost-efficiency without surrendering competitiveness.} The curated model sits on the cost-efficient Pareto front against six of seven length-comparable external models; the lone exception trades accuracy away for its cost edge. The eighth external model (Qwen3.5-4B) admits no length-controlled comparison at all: it is never observed at a concise operating point.
\end{takeaway}

\subsection{Verbose tokens earn their cost on a shrinking window of capabilities}
\label{sec:verbosity}

\S\ref{sec:cost}--\S\ref{sec:frontier} establish that the curated model is competitive on accuracy while emitting dramatically fewer tokens. The natural follow-up is the inverse question: \emph{when are verbose responses actually contributing to correctness, and when are they filler?} We answer it on a \textbf{triple dissociation} across three conditions in the pool: the \curated model (\nDat, \cD) is concise and non-reasoning (${\sim}30$ tokens per response); \nMam (\cM), the uncurated MAmmoTH-VL baseline, is verbose (${\sim}60$ tokens) but non-reasoning; and \nQwen (\cQ) is both verbose (170 tokens at 2B, 1{,}284 at 4B) and reasoning-structured (chain-of-thought). We refer to the three by these initials throughout.

Our operational definition of ``verbose tokens don't contribute anything'' is: on the same example, $\cQ \approx \cD$ on correctness. The verbose model produces no per-example accuracy advantage over the concise model. The dissociation distinguishes whether verbosity \emph{as such} is the lever (\cM would beat \cD) or whether \emph{reasoning-structured} verbosity is doing the work (\cQ would beat \cD and \cM, \cM $\approx$ \cD). 

We run the analysis on all 18 evaluations of the grouped frontier suite at the two scales where the dissociation is available: \textbf{2B} (\nDat 2B, \nMam 2B, \nQwen-2B) and \textbf{4B} (\nDat 4B, \nMam 4B, \nQwen-4B). How we binarize per-example correctness, define the pairwise deltas ($\cQ-\cD$, $\cM-\cD$, $\cQ-\cM$), and assign each (scale, capability) comparison a regime is set out in \S\ref{subsec:m-dissociation}; the per-example contingency machinery is in Appendix~\ref{app:tablea1}.

\paragraph{Generic verbosity (Mammoth) contributes nothing, anywhere.}
\nMam's verbose-but-not-reasoning outputs never show a positive accuracy advantage over the \curated model at any capability or scale: $\cM-\cD$ is negative everywhere, from $-2.6$~pp at best (2B Captioning and Chart \& Diagram) to $-54.9$~pp at worst (4B Referring \& Grounding, where Mammoth largely fails to emit valid bounding boxes). This is the cleanest empirical version of the ``tokens not paid for'' claim: a model emitting 2--3$\times$ the \curated baseline's tokens, without structuring them as reasoning, recovers none of the accuracy gap and often opens new ones. \textbf{Generic verbosity does not buy correctness.}

\begin{figure}[h]
  \centering
  \includegraphics[width=\linewidth]{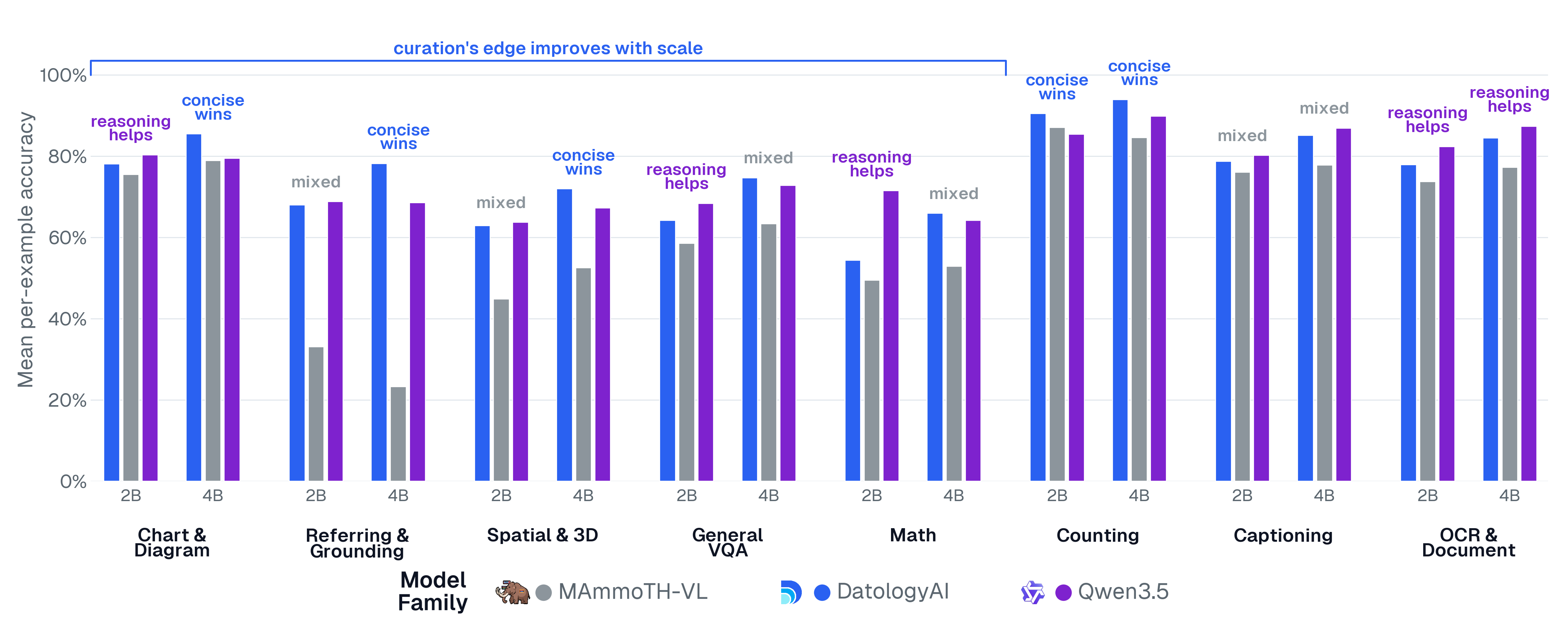}
  \caption{\textbf{Verbose reasoning loses its accuracy advantage at scale.} Each capability pairs its 2B cluster with its 4B cluster, every cluster a trio of DatologyAI ($D$), Mammoth ($M$), and Qwen3.5 ($Q$) mean accuracy, so the scale shift reads within a capability. The label above each cluster names the regime, assigned by comparing the \curated model against each verbose model at a 2-percentage-point threshold: \textcolor[HTML]{2A61F1}{\emph{concise wins}} when $D$ beats both $Q$ and $M$; \textcolor[HTML]{7E22CE}{\emph{reasoning helps}} when $Q$ beats $D$ while the non-reasoning $M$ trails it (reasoning-structured verbosity earns its tokens where generic verbosity does not); and \textcolor[HTML]{8C959C}{\emph{mixed}} otherwise. Capabilities are ordered left-to-right by how that regime moves with scale; the denoted region marks the five where the \curated model gains ground from 2B to 4B. Reasoning's winning window shrinks from four capabilities at 2B to one (OCR \& Document) at 4B.}
  \label{fig:verbosity-bars}
\end{figure}

\paragraph{Reasoning-structured verbosity helps on a window that shrinks with scale.}
\nQwen's chain-of-thought verbosity is a different story, but a shrinking one. At 2B it beats the \curated model by $\geq 2$~pp on four of eight capability groups (led by Math, $+17.1$~pp); at 4B that window collapses to one (OCR \& Document, $+2.9$~pp), and on four groups the concise \curated model instead wins by $\geq 2$~pp, by as much as $9.6$~pp on Referring \& Grounding (\Cref{fig:verbosity-bars}; full table in \Cref{tab:dissociation-main}). The shrink is \emph{concordant} with \S\ref{sec:quality}'s length-controlled regression, where the curation effect grows super-linearly with scale ($+15.6$~pp at 2B, $+21.2$~pp at 4B); the capability-level attribution this section adds leaves OCR \& Document as the lone domain where reasoning still pays its way (structured step-by-step transcription benefits from explicit intermediate steps).

\paragraph{Within-Captioning divergence: CAPability and DetailCaps point in opposite directions.}
The Captioning row aggregates two evals with very different demands, and they point in \emph{opposite} directions at both scales (\Cref{tab:captioning-split}):

\begin{table}[h]
  \centering
  \small
  \begin{tabular}{l l r r r r r}
    \toprule
    Scale & Eval & \cD & \cM & \cQ & $\cQ{-}\cD$ & $\cM{-}\cD$ \\
    \midrule
    \textbf{2B} & CAPability (object recall)                  & 0.672 & 0.669 & 0.803 & \textbf{+13.1} & $-0.3$ \\
    \textbf{2B} & DetailCaps (caption precision, thr.\ 0.7)   & 0.901 & 0.852 & 0.799 & \textbf{\boldmath$-10.2$} & $-4.9$ \\
    \textbf{4B} & CAPability (object recall)                  & 0.738 & 0.681 & 0.805 & \textbf{+6.7}  & $-5.6$ \\
    \textbf{4B} & DetailCaps (caption precision, thr.\ 0.7)   & 0.964 & 0.874 & 0.932 & \textbf{\boldmath$-3.2$} & $-9.1$ \\
    \bottomrule
  \end{tabular}
  \caption{\textbf{Within-Captioning divergence.} CAPability (recall) rewards verbose enumeration; DetailCaps (precision) penalizes it. The group's ``Mixed'' verdict averages a recall gain and a precision loss.}
  \label{tab:captioning-split}
\end{table}

CAPability tests whether a caption \emph{mentions} a set of ground-truth objects: a recall-style task where verbose enumeration helps, because more tokens are more chances to name the right object. DetailCaps tests whether the claims a caption \emph{makes} are supported by reference captions: a precision-style task where verbose elaboration can hurt, because each additional claim is another opportunity for an unsupported assertion. The Captioning group's net ``Mixed / inconclusive'' verdict at both scales averages a recall gain ($+13.1$ / $+6.7$) and a precision loss ($-10.2$ / $-3.2$). Treating ``Captioning'' as a single capability obscures this. A reader interested in caption \emph{recall} should expect verbose reasoning to help; a reader interested in caption \emph{precision} should not.

\paragraph{Per-example contingency: \cQ and \cD have different error sets.}
Mean accuracy says which model is more often correct; it hides \emph{how} the two disagree. A per-example contingency on shared examples shows that \textbf{\cQ and \cD reach substantially different answers across nearly every capability}: at 4B, \nQwen's rescue rate (the fraction of \cD's misses it gets right) exceeds $25\%$ in every capability, and beats its own misled rate on 7 of 8. Yet mean accuracy still favors \cD on most of them, because when \cD's accuracy floor is high, a small misled rate on the large ``\cD-correct'' base outweighs a large rescue rate on the small ``\cD-wrong'' base. The two models are genuinely complementary at the example level, but the concise model still ships more correct answers per query; the contingency view is the lens that matters for routing, ensembling, or conditional reasoning, not for the cost-per-correct framing of \S\ref{sec:intro} and \S\ref{sec:conclusion}. The full rates, the reconciling identity, and a worked example are in Appendix~\ref{app:tablea1}.

\paragraph{Benchmarks that diverge from their capability group.}
Five individual (scale, benchmark) cells sit $>10$~pp away from their capability-group mean. Two are the CAPability / DetailCaps split discussed above; the other three are PixMo Points at both scales and OCRBench at 2B:
\begin{itemize}
  \item \textbf{PixMo Points} (Referring \& Grounding): $\cQ-\cD = -24.6$~pp at 4B and $-9.6$~pp at 2B, vs the group means of $-9.6$~pp and $+0.8$~pp respectively. The pointing-output task format (predicting structured \texttt{<point x="..." y="...">} tags) is uniquely hostile to chain-of-thought reasoning models, which appear to talk themselves out of correct coordinates: the $\cQ_{\text{misled}}$ rate at 4B is \textbf{73.4\%}, the highest of any (scale, benchmark) pair. This is consistent with \S\ref{sec:frontier}'s note that grounding tasks are exactly where the \curated model's concision is operationally most valuable.
  \item \textbf{OCRBench at 2B}: $\cQ-\cD = +15.7$~pp vs the OCR \& Document group mean of $+4.5$~pp. OCRBench's free-form character-recognition style benefits from explicit reasoning more than the document-QA tasks (DocVQA, TextVQA) in the same group.
\end{itemize}
Full per-benchmark numbers (mean accuracy and conditional rates) are in the appendix (Table~\ref{tab:a1}).

\begin{takeaway}
\textbf{Verbose tokens earn their cost on a capability window that shrinks with scale.} Generic, non-reasoning verbosity (\nMam's $2$--$3\times$ token count) helps nowhere: $\cM-\cD$ is negative for every capability at both scales. Reasoning-structured verbosity (\nQwen's chain-of-thought) helps 4 of 8 capability groups at 2B but only 1 at 4B, where the concise \curated model instead wins 4 of 8. Per example, \cQ and the \curated model have substantially different error sets (rescue rate $\geq 25\%$ everywhere), yet \cD's high accuracy floor keeps most of that from moving mean accuracy in \cQ's favor. At scale, brevity-via-curation appears to \emph{remove} the conditions under which verbose reasoning would have paid.
\end{takeaway}

\section{Conclusion}
\label{sec:conclusion}

This work argued for \textbf{inference efficiency through brevity, achieved through data curation}, with our VLM pretraining curation as the case study:

\paragraph{What the analyses establish.}

\noindent\textbf{\boldmath (1) Brevity saves money, by $35\times$ per correct answer at 4B.} Datology 4B produces a correct answer for 0.41 TFLOPs to Qwen3.5-4B's 14.58, a $35\times$ Cost-of-Pass gap within ${\sim}1$~pp of accuracy (0.691 vs 0.704); within the matched-scale curated-vs-uncurated pairs, the curated model costs $2.2\times$ / $1.6\times$ / $2.3\times$ fewer FLOPs per correct answer at 1B / 2B / 4B (\S\ref{sec:cost}).

\noindent\textbf{(2) Quality holds at matched length, and the effect grows with scale.} At the same output length on the same benchmark, the curated model is $+17.55$~pp more accurate than its uncurated pair (pooled across scales), and the per-scale effect rises from $+15.6$~pp at 2B to $+21.2$~pp at 4B: curation pays off more at larger scale, not less (\S\ref{sec:quality}).

\noindent\textbf{(3) Generic verbosity buys nothing, and reasoning verbosity's edge narrows with scale.} The verbose non-reasoning model never beats the curated model on any of the 16 scale $\times$ capability cells, and reasoning-structured verbosity's edge narrows from 4 of 8 capability groups at 2B to 1 of 8 at 4B (\S\ref{sec:verbosity}).

\noindent\textbf{(4) Reasoning is a distinct route, not a cheaply-emulable one.} Per example, the curated model reaches correct answers the verbose reasoning model misses, and vice versa. This suggests that curating for reasoning and curating for brevity are complementary levers, and that a model trained to be both concise and a stronger reasoner is the natural next build (\S\ref{sec:verbosity}).

\paragraph{Takeaway.}
We argue for inference efficiency through brevity, achieved through data curation. The standard toolkit (quantization, distillation, MoE, speculative decoding) works on the per-token cost and treats output length as a fixed input. Output length is the bigger lever in practice, and curation pulls it without a quality tax: at the scales we measured, the same curation move yields shorter outputs \emph{and} higher matched-length accuracy, the advantage reaching a $35\times$ Cost-of-Pass gap at 4B and the accuracy effect growing super-linearly in model size.

Per-token serving cost keeps falling through hardware and software gains, but per-correct cost is not following it down, because cheaper tokens \emph{induce} more demand (the Jevons rebound) on two fronts: longer reasoning chains within a query, and more queries across users. The within-query front is already visible across the current open-weight 2B--4B VLM pool, where mean output length spans roughly two orders of magnitude and the most expensive comparators by Cost-of-Pass are systematically the most verbose (\S\ref{sec:cost}). Brevity-per-correct removes that front: holding quality per task fixed pins the within-query token count, leaving demand induction with one front instead of two. Inference efficiency in this regime is not a per-token-cost problem; it is a \textbf{tokens-per-correct} problem, and brevity is the lever that targets it directly. Curation is how you pull it, and the reason it works is amortization: the cost of concision is paid once, at training, and compiled into the weights rather than re-paid at every decode. That points forward, too. Production usage can reveal where a deployed model still over-spends tokens; each such pattern is a target for the next curation pass; and the loop tightens with every iteration, until amortized inference becomes continual learning.

Pascal apologized for the long letter because he had no time to make it shorter. Brevity is work, and a model trained on verbose data is in Pascal's position at every generation, with no time to shorten any answer. There are two places to find that time: decode-time methods buy it back one query at a time, shortening each letter as it is written, while curation buys it once, before the model ships, with the saving amortized across every answer the model will ever produce. Pascal paid for brevity by the letter; curation pays for it by the corpus, and collects on every generation thereafter.

\clearpage
\section{Contributions and Acknowledgements}
\label{sec:contrib}

\begin{tabularx}{\textwidth}{@{}p{0.19\textwidth}X@{}}
\textbf{Core Contributors} & Matthew Leavitt, Sid Joshi \\[0.25em]
& \emph{for making this letter longer than usual.} \\
\noalign{\vspace{0.75em}}

\textbf{Technical Contributors} & Haoli Yin, Rishabh Adiga, Haakon Mongstad, and Alvin Deng, \\
\noalign{\vspace{0.25em}}
& \emph{the giants, whose shoulders proved eminently standable.}\\
\noalign{\vspace{0.75em}}

\textbf{Leadership} & David Schwab, Bogdan Gaza, Ari Morcos, \\[0.25em]
& \emph{for watching the horizon while the rest of us watched the loss curves.} \\
\noalign{\vspace{0.75em}}

\textbf{Acknowledgements} & Kaleigh Mentzer, Luke Merrick, and Pratyush Maini, \\
\noalign{\vspace{0.25em}}
& \emph{for thoughtful feedback that made the work shorter AND better.}
\end{tabularx}

\bibliographystyle{abbrvnat}
\clearpage
\bibliography{references}

\begin{thebibliography}{25}
\providecommand{\natexlab}[1]{#1}
\providecommand{\url}[1]{\texttt{#1}}
\expandafter\ifx\csname urlstyle\endcsname\relax
  \providecommand{\doi}[1]{doi: #1}\else
  \providecommand{\doi}{doi: \begingroup \urlstyle{rm}\Url}\fi

\bibitem[Brown(2026)]{brown2026}
N.~Brown.
\newblock Report capability against a compute budget, not as a single number.
\newblock Post on X (OpenAI), 2026.
\newblock URL \url{https://x.com/polynoamial/status/2064210146558136827}.

\bibitem[Chen et~al.(2024)]{chen2024mllmjudge}
D.~Chen et~al.
\newblock {MLLM}-as-a-judge: Assessing multimodal {LLM}-as-a-judge with
  vision-language benchmark.
\newblock \emph{arXiv preprint arXiv:2402.04788}, 2024.
\newblock URL \url{https://arxiv.org/abs/2402.04788}.

\bibitem[DatologyAI(2026)]{datology2026vlm}
DatologyAI.
\newblock 20/20 vision language models: A prescription for better {VLMs}
  through data curation alone.
\newblock \emph{arXiv preprint arXiv:2605.11405}, 2026.
\newblock URL \url{https://arxiv.org/abs/2605.11405}.

\bibitem[Du et~al.(2025)Du, Kang, Han, Krishna, and Zhu]{ockbench2025}
Z.~Du, H.~Kang, S.~Han, T.~Krishna, and L.~Zhu.
\newblock {OckBench}: Measuring the efficiency of {LLM} reasoning.
\newblock \emph{arXiv preprint arXiv:2511.05722}, 2025.
\newblock URL \url{https://arxiv.org/abs/2511.05722}.

\bibitem[Dubois et~al.(2024)Dubois, Galambosi, Liang, and
  Hashimoto]{dubois2024length}
Y.~Dubois, B.~Galambosi, P.~Liang, and T.~B. Hashimoto.
\newblock Length-controlled {AlpacaEval}: A simple way to debias automatic
  evaluators.
\newblock \emph{arXiv preprint arXiv:2404.04475}, 2024.
\newblock URL \url{https://arxiv.org/abs/2404.04475}.

\bibitem[{Epoch AI}(2025)]{epoch2025length}
{Epoch AI}.
\newblock {LLM} responses are getting longer.
\newblock Epoch AI Data Insight, 2025.
\newblock URL \url{https://epoch.ai/data-insights/output-length}.

\bibitem[Erol et~al.(2025)]{erol2025costofpass}
M.~H. Erol et~al.
\newblock Cost-of-pass: An economic framework for evaluating language models.
\newblock \emph{arXiv preprint arXiv:2504.13359}, 2025.
\newblock URL \url{https://arxiv.org/abs/2504.13359}.
\newblock Stanford.

\bibitem[Gershman and Goodman(2014)]{gershman2014amortized}
S.~J. Gershman and N.~D. Goodman.
\newblock Amortized inference in probabilistic reasoning.
\newblock In \emph{Proceedings of the 36th Annual Conference of the Cognitive
  Science Society (CogSci)}, 2014.
\newblock URL \url{https://dblp.org/rec/conf/cogsci/GershmanG14.html}.

\bibitem[Gundlach et~al.(2025)Gundlach, Lynch, Mertens, and
  Thompson]{priceofprogress2025}
H.~Gundlach, J.~Lynch, M.~Mertens, and N.~Thompson.
\newblock The price of progress: Price performance and the future of {AI}.
\newblock \emph{arXiv preprint arXiv:2511.23455}, 2025.
\newblock URL \url{https://arxiv.org/abs/2511.23455}.

\bibitem[Guo et~al.(2024)]{guo2024mammothvl}
J.~Guo et~al.
\newblock {MAmmoTH-VL}: Eliciting multimodal reasoning with instruction tuning
  at scale.
\newblock \emph{arXiv preprint arXiv:2412.05237}, 2024.
\newblock URL \url{https://arxiv.org/abs/2412.05237}.

\bibitem[Holtzman et~al.(2020)Holtzman, Buys, Du, Forbes, and
  Choi]{holtzman2020curious}
A.~Holtzman, J.~Buys, L.~Du, M.~Forbes, and Y.~Choi.
\newblock The curious case of neural text degeneration.
\newblock In \emph{International Conference on Learning Representations
  (ICLR)}, 2020.
\newblock URL \url{https://arxiv.org/abs/1904.09751}.

\bibitem[Huang et~al.(2024)Huang, Qiu, Wang, Ponti, and Titov]{huang2024length}
Z.~Huang, Z.~Qiu, Z.~Wang, E.~M. Ponti, and I.~Titov.
\newblock Post-hoc reward calibration: A case study on length bias.
\newblock \emph{arXiv preprint arXiv:2409.17407}, 2024.
\newblock URL \url{https://arxiv.org/abs/2409.17407}.

\bibitem[Jung et~al.(2025)Jung, Lee, Kim, and Yoon]{jung2025visual}
M.~Jung, S.~Lee, E.~Kim, and S.~Yoon.
\newblock Visual attention never fades: Selective progressive attention
  recalibration for detailed image captioning in multimodal large language
  models.
\newblock \emph{arXiv preprint arXiv:2502.01419}, 2025.
\newblock URL \url{https://arxiv.org/abs/2502.01419}.

\bibitem[Kaiser et~al.(2025)Kaiser, Frigessi, Ramezani-Kebrya, and
  Ricaud]{cogniload2025}
D.~Kaiser, A.~Frigessi, A.~Ramezani-Kebrya, and B.~Ricaud.
\newblock {CogniLoad}: A synthetic natural language reasoning benchmark with
  tunable length, intrinsic difficulty, and distractor density.
\newblock \emph{arXiv preprint arXiv:2509.18458}, 2025.
\newblock URL \url{https://arxiv.org/abs/2509.18458}.

\bibitem[Kaiser et~al.(2026)Kaiser, Frigessi, Ramezani-Kebrya, and
  Ricaud]{decomposing2026}
D.~Kaiser, A.~Frigessi, A.~Ramezani-Kebrya, and B.~Ricaud.
\newblock Beyond accuracy: Decomposing the reasoning efficiency of {LLMs}.
\newblock \emph{arXiv preprint arXiv:2602.09805}, 2026.
\newblock URL \url{https://arxiv.org/abs/2602.09805}.

\bibitem[Lee et~al.(2025)Lee, Yoon, Bui, Shi, and Yoon]{lee2025captioning}
S.~Lee, S.~Yoon, T.~Bui, J.~Shi, and S.~Yoon.
\newblock Enhancing factuality in detailed image captioning with {LLM}-{MLLM}
  collaboration.
\newblock In \emph{International Conference on Learning Representations
  (ICLR)}, 2025.
\newblock URL \url{https://openreview.net/forum?id=psIymxANmd}.
\newblock OpenReview psIymxANmd.

\bibitem[Murray and Chiang(2018)]{murray2018correcting}
K.~Murray and D.~Chiang.
\newblock Correcting length bias in neural machine translation.
\newblock In \emph{Proceedings of the Third Conference on Machine Translation
  (WMT)}, 2018.
\newblock URL \url{https://arxiv.org/abs/1808.10006}.

\bibitem[{Nous Research}(2025)]{nous2025thinking}
{Nous Research}.
\newblock Measuring thinking efficiency in reasoning models: The missing
  benchmark.
\newblock Nous Research, 2025.
\newblock URL
  \url{https://nousresearch.com/measuring-thinking-efficiency-in-reasoning-models-the-missing-benchmark}.

\bibitem[{OpenAI}(2026)]{openaiReasoningEffort}
{OpenAI}.
\newblock Reasoning effort parameter.
\newblock OpenAI API documentation, 2026.
\newblock URL \url{https://developers.openai.com/api/docs/guides/reasoning}.

\bibitem[Park et~al.(2024)Park, Rafailov, Ermon, and
  Finn]{park2024disentangling}
R.~Park, R.~Rafailov, S.~Ermon, and C.~Finn.
\newblock Disentangling length from quality in direct preference optimization.
\newblock \emph{arXiv preprint arXiv:2403.19159}, 2024.
\newblock URL \url{https://arxiv.org/abs/2403.19159}.

\bibitem[Pichai(2026)]{pichai2026io}
S.~Pichai.
\newblock Google i/o 2026 keynote: token-volume growth.
\newblock Google Blog, 2026.
\newblock URL
  \url{https://blog.google/innovation-and-ai/sundar-pichai-io-2026/}.

\bibitem[Singhal et~al.(2024)Singhal, Goyal, Xu, and
  Durrett]{singhal2024length}
P.~Singhal, T.~Goyal, J.~Xu, and G.~Durrett.
\newblock A long way to go: Investigating length correlations in {RLHF}.
\newblock \emph{Conference on Language Modeling (COLM)}, 2024.
\newblock URL \url{https://arxiv.org/abs/2310.03716}.
\newblock arXiv:2310.03716.

\bibitem[Sutton(2019)]{sutton2019bitter}
R.~S. Sutton.
\newblock The bitter lesson, 2019.
\newblock URL \url{http://www.incompleteideas.net/IncIdeas/BitterLesson.html}.

\bibitem[Welleck et~al.(2020)Welleck, Kulikov, Roller, Dinan, Cho, and
  Weston]{welleck2020neural}
S.~Welleck, I.~Kulikov, S.~Roller, E.~Dinan, K.~Cho, and J.~Weston.
\newblock Neural text generation with unlikelihood training.
\newblock In \emph{International Conference on Learning Representations
  (ICLR)}, 2020.
\newblock URL \url{https://arxiv.org/abs/1908.04319}.

\bibitem[Wu et~al.(2016)Wu, Schuster, Chen, Le, Norouzi, et~al.]{wu2016google}
Y.~Wu, M.~Schuster, Z.~Chen, Q.~V. Le, M.~Norouzi, et~al.
\newblock Google's neural machine translation system: Bridging the gap between
  human and machine translation.
\newblock 2016.
\newblock URL \url{https://arxiv.org/abs/1609.08144}.

\end{thebibliography}

\beginappendix
\numberwithin{figure}{section}
\numberwithin{table}{section}

\renewcommand{\thesection}{M}
\section{Methodology}
\label{app:methods}

\subsection{The model pool \& eval grid}
\label{subsec:m-pool}

The model pool is built to do two things at once: matched-scale controlled pairs that isolate the curation effect at each size, and a set of frontier comparators that place it in context. We include 14 models:
\begin{itemize}
  \item \textbf{Datology} and \textbf{Mammoth} at 1B, 2B, and 4B activated parameters (six internal models), the controlled matched-scale comparison. At each scale the pair shares an identical backbone, VL recipe, and 4k context; only the pretraining data differs. Each matched pair isolates the \emph{curation} effect from confounds of architecture or training procedure, and spanning three scales lets us measure how that effect varies with model size.
  \item 8 frontier open VLMs at 2--4B activated parameters: Qwen3.5-2B, Qwen3.5-4B, Qwen3-VL-2B-Instruct, Qwen3-VL-4B-Instruct, InternVL3-2B, InternVL3.5-2B, InternVL3.5-4B, Perceptron-Isaac-0.2-2B-Preview.
\end{itemize}
For internal models with multiple training seeds, seeds are pooled at the per-example level before any per-model aggregate is computed: seed-average the per-example signal, then aggregate to per-model. This keeps the per-model statistic conservative, since variance across seeds shows up as wider example-level distributions rather than an inflated effective sample size.

The evaluation grid is 20 frontier evals across 8 capability families: Referring \& Grounding, General VQA, OCR \& Document, Captioning, Spatial \& 3D, Counting, Chart \& Diagram, and Math (BLINK and BrandID are aggregate-only and appear only at the highest level of aggregation). Each eval has a single headline metric (recall@0.5 for COCO-like grounding tasks, accuracy for MCQ, CAPTURE-MicroOverall for DetailCaps, F1-MicroOverall for CAPability, normalized OCRBench, and so on), normalized to $[0, 1]$.

\subsection{Inference cost model}
\label{subsec:m-cost}

The headline cost numbers in this post come from a decode-dominated FLOPs proxy,
\begin{equation*}
  \text{FLOPs}_{\text{response}} \approx 2N\,n_{\text{out}},
\end{equation*}
where $N$ is active parameters and $n_{\text{out}}$ is the number of generated tokens. We use \textbf{mean} generated tokens per response as the central tendency throughout, for two reasons. Mean is the right quantity for cost forecasting, since hosting bills sum over tokens, not percentiles. And it is the central tendency that makes the inference-cost claim concrete: ``this model costs \$X per million correct answers'', not ``this model sits in some percentile of cost-efficiency''.

\paragraph{Scope of the proxy.} The $2N\,n_{\text{out}}$ term is the decode feed-forward cost, linear in active parameters and output length. It omits attention, which is quadratic in sequence length, and it assumes compute scales with active parameters, setting aside the bandwidth and memory-capacity limits that dominate for larger or sparsely-activated models. For the small dense VLMs we study, with responses of tens of tokens, both omissions are negligible and the proxy is accurate; for the long-output reasoning models they are not. Attention is a non-trivial share of Qwen3.5-4B's ${\sim}1{,}300$-token responses, so the proxy \emph{understates} their cost, and because the understatement grows with output length, every Cost-of-Pass gap we report between a concise model and a verbose one is conservative.

\paragraph{Two pricing lanes.}
\begin{itemize}
  \item \emph{Provider-hosted (for frontier open models):} dollar cost per output token is the average of Together and Fireworks public per-token rates as of the data-collection date, fixed per-model.
  \item \emph{Self-hosted (for Datology and a head-to-head appendix on the frontier set):} a memory-bandwidth roofline,
  \begin{equation*}
    \text{cost/token} = \frac{\$2.5/\text{h}}{3600} \cdot \frac{b \cdot 2N}{b \cdot \text{BW}} = \frac{\$2.5/\text{h} \cdot 2N}{3600 \cdot \text{BW}},
  \end{equation*}
  with $\text{BW} = 3.35$~TB/s (H100) and 2 bytes/param (FP16). Decode is memory-bandwidth-bound: all sequences in a batch share a single weight-loading pass through HBM, so wall-clock time per token is approximately independent of batch within the bandwidth bound, but that wall-clock cost amortizes across the $B$ sequences, so \textbf{\boldmath per-token dollar cost scales as $1/B$}. We use \textbf{\boldmath$B = 64$} as a log-anchored placeholder; the $1/B$ factor is baked into all CoP figures in this post, and cost \emph{ratios} between models are batch-invariant.
\end{itemize}
From these we derive \textbf{Cost-of-Pass}, the dollar cost per correct answer on a given eval:
\begin{equation*}
  \text{CoP} = \frac{(\text{cost-per-output-token}) \cdot n_{\text{out}}}{\text{accuracy}}.
\end{equation*}
CoP is the central cost number in the post: it has the right dimensional analysis to support the \emph{brevity saves you money} framing.

\paragraph{Dollar Cost-of-Pass, worked.} Instantiating CoP under the self-hosted roofline (\$2.5/hr, $\text{BW} = 3.35$~TB/s, FP16, $B = 64$),
\begin{equation*}
  \frac{\$}{\text{M correct}} = \frac{\$2.5/3600}{B \cdot \text{BW}} \cdot 2N \cdot \frac{\bar{n}_{\text{out}}}{\text{accuracy}}.
\end{equation*}
a million correct answers cost \$0.33 from Datology 1B, \$0.87 from Datology 2B, \$1.34 from Datology 4B, \$3.16 from Qwen3.5-2B, and \$47.24 from Qwen3.5-4B: the same ${\sim}40\times$ ratio between the \curated 4B and Qwen3.5-4B that the FLOPs accounting reports, plus a ${\sim}10\times$ ratio between the \curated 1B and Qwen3.5-2B at comparable accuracy. A million correct answers from Datology 4B costs roughly the price of a soda; from Qwen3.5-4B, roughly a tank of gas. The dollar cost is linear in FLOPs/correct at fixed batch and bandwidth, so the dollar ratios equal the FLOPs ratios and the headline gaps are invariant to reasonable hardware assumptions.

\subsection{Quality measurement: what we test, what we don't claim}
\label{subsec:m-quality}

The brevity-saves-money claim has two halves: brevity reduces cost (\S\ref{subsec:m-cost}), and brevity doesn't sacrifice the things you'd worry about losing. The second half needs a quality measurement that is \emph{fair across length distributions}, so verbose models aren't penalized for verbosity itself, and that \emph{separates the dimensions that matter}: a model that is concise but omits things is failing differently from one that is concise and grounded.

We instrument two captioning evaluations from the suite: \textbf{DetailCaps} (precision-side: did the model assert things that are correct?) and \textbf{CAPability} (recall-side: did the model cover the things it should?). They're chosen because they're the captioning evals in our frontier set with per-example signals we can manipulate, and because length is the natural axis they vary along.

\paragraph{Boundary of inquiry.} All our quality signals in this section are \emph{text-based}: model caption against reference captions (DetailCaps precision) or against a single ground-truth object label (CAPability recall). We are explicitly \emph{not} doing image-grounded measurement here: the judge does not see the image. Where this changes how a result should be read, we caveat. See \S\ref{subsec:m-judge} for the long discussion of what this rules in and out for the hallucination claim specifically.

\subsection{Why lexical-overlap precision is the wrong tool, and what an LLM judge does instead}
\label{subsec:m-judge}

We tried two precision measurements on DetailCaps and got incompatible answers. Working out which one to trust is itself the story.

\paragraph{Method 1: lexical-overlap object precision.} For each (caption, references) pair: extract content lemmas from the caption (lowercase alphabetic tokens $\geq 3$ chars, drop stopwords and caption-boilerplate, light plural-to-singular fold), do the same for the union of the three references, compute precision $= |\text{mentioned} \cap \text{supported}| / |\text{mentioned}|$. This is a fast deterministic proxy for CAPTURE-style scoring that doesn't need the CAPTURE dependency stack. It's also the standard cheap object-precision proxy that shows up in a lot of hallucination tooling.

\paragraph{Method 2: a frontier LLM as a claim-level judge.} For each (caption, references) pair: ask the judge (Claude Haiku 4.5, temperature 0) to decompose the caption into \emph{atomic factual claims}, then label each as one of three:
\begin{itemize}
  \item \textbf{supported}: some reference confirms or paraphrases it semantically (lexical match not required).
  \item \textbf{unsupported}: some reference contradicts it, or the candidate asserts the existence or identity of an object or attribute that no reference mentions and that should reasonably be visible. \emph{This is the hallucination signal.}
  \item \textbf{unverifiable}: the claim is about something the references genuinely do not address (subjective mood, emotion, intent, abstract interpretation, off-camera inference, photographer style). \textbf{Excluded from the precision denominator entirely.}
\end{itemize}
Precision $= \text{supported} / (\text{supported} + \text{unsupported})$. Temperature 0, deterministic across re-runs. The prompt was smoke-validated by hand against actual images for 200 calls before committing to the full batch.

\paragraph{The headline disagreement.}
Across all 14 models on the full 4,870 DetailCaps examples (68,180 judge calls in the full batch), \Cref{tab:judge-disagreement} reports the split:

\begin{table}[h]
  \centering
  \small
  \begin{tabular}{l r r}
    \toprule
    Metric & Lexical proxy & LLM judge \\
    \midrule
    Cross-model Spearman(length, hallucination\_rate)  & \textbf{\boldmath$+0.94$} & \textbf{\boldmath$-0.018$} \\
    Per-model precision range                          & 0.31 -- 0.64 & 0.77 -- 0.88 \\
    Cross-model rank agreement (LLM vs lex Spearman)   & --- & \textbf{\boldmath$-0.018$} \\
    \bottomrule
  \end{tabular}
  \caption{\textbf{Lexical and LLM-judge precision disagree on which models hallucinate.} The two methods even rank Qwen3.5-4B oppositely (worst under lexical, best under the judge).}
  \label{tab:judge-disagreement}
\end{table}

The cross-model rank-agreement number is the one that decides it: the two methods disagree on \emph{which models are more precise}, not just on what ``precise'' numerically means. They rank Qwen3.5-4B as the \textbf{worst} (lexical, precision 0.313) and the \textbf{best} (LLM, precision 0.879) of the 14. That's not a calibration disagreement; that's a measurement disagreement.

\paragraph{Worked example: Qwen3.5-4B on a typical caption.}
Qwen3.5-4B writes ${\sim}5{,}200$-character captions with structured reasoning preambles (``The user wants a detailed description\ldots\ 1. Identify the main subject: \ldots\ 2. Analyze the central figure: \ldots''). On the Monopoly-Man graffiti image we hand-inspected in the smoke review:
\begin{itemize}
  \item \textbf{Lexical method} counts every content lemma. A 5,200-char caption produces ${\sim}600$+ content lemmas after stopword filtering, including reasoning scaffolding (``identify'', ``subject'', ``central'', ``figure'', ``analysis'') and meta-commentary (``appears'', ``depicts'', ``rendered''). Many of these aren't in the human references, so they count as unsupported by the denominator-inflation mechanism. The proxy returns precision $\approx 0.30$ for this caption.
  \item \textbf{LLM judge} extracts ${\sim}20$ atomic factual claims: ``the artwork is graffiti on a wall'' (supported), ``the central figure is a Monopoly Man'' (supported), ``to the left is a vertical wooden beam'' (unsupported, references don't mention a beam), ``the style is reminiscent of 1980s graffiti'' (unverifiable, interpretive). It ignores the scaffolding because it isn't a factual claim. Judge returns precision $= 0.85$.
\end{itemize}

\paragraph{Why the mechanisms diverge.} Both methods compute supported / total, but they define \emph{total} differently:
\begin{itemize}
  \item \textbf{Lexical ``total''} $=$ bag of content lemmas, which scales roughly linearly with caption length. Adding structure, hedge words, or reasoning prose inflates the denominator without contributing supportable claims.
  \item \textbf{LLM ``total''} $=$ number of atomic factual assertions, which scales much more slowly with length. In our data, Qwen3.5-4B's 5,229-char captions average 18 claims; InternVL3.5-4B's 752-char captions average 13.3 claims: $7\times$ the characters, $1.35\times$ the claims.
\end{itemize}
Length inflates the lexical denominator faster than the supportable numerator can keep up, mechanically lowering precision for verbose models. \textbf{\boldmath This fully explains the $+0.94$ cross-model length-vs-hallucination correlation under the lexical method: it's a length-counting artifact, not a hallucination signal.}

\paragraph{Why we trust the LLM-judge result for cross-model claims.} Despite known LLM-judge limitations:
\begin{enumerate}
  \item \textbf{Smoke-validated against actual images.} We hand-checked 200 calls and their per-claim verdicts against the source images and confirmed the judge correctly catches misidentifications, OCR errors, and false attribute claims; correctly routes subjective material to \emph{unverifiable}; rarely over-decomposes; and rarely fabricates claims not in the candidate (${\sim}1$ minor instance in 200 calls).
  \item \textbf{Temperature 0}, deterministic, no reproducibility drift between calls or re-runs.
  \item \textbf{The \emph{unverifiable} bucket explicitly separates \emph{what the references don't address} from \emph{what the candidate got wrong}.} The lexical proxy cannot make this distinction; any candidate content not in the references is forced into the unsupported pile regardless of whether it's a hallucination or simply outside the references' scope.
\end{enumerate}

\paragraph{Where we drew the boundary of inquiry.} Both measurements compare \emph{caption text to reference text}. Neither sees the image. We're claiming that \emph{under the operational definition of ground truth used by the DetailCaps benchmark, that the three human reference captions are an accurate description of the image, the LLM judge gives cross-model precision scores that don't track length.} What would invalidate that claim is the references themselves diverging from the image: a reference says ``yellow car'' when the image contains a blue one, or omits a clearly-visible elephant. Verifying reference fidelity is exactly what an image-grounded judge would test, and we treat it as out of scope here because the benchmark-quality question (``are DetailCaps' references a faithful representation of the image?'') and the brevity question (``does verbosity buy precision in modern VLMs?'') are conceptually separable, with an image-grounded judge answering the former.

An image-grounded study, canonical CHAIR with COCO instance masks or a VLM judge that sees the image, could give a different cross-model length-vs-precision number. We don't pre-empt that and we don't claim to overturn image-grounded prior literature on length and hallucination. The honest position for a reader: if you have prior reasons to doubt DetailCaps' references as a proxy for image content, the right place to discount our result is the benchmark-design discussion, not the brevity-vs-precision finding.

\subsection{Why we don't use the existing CAPability judge for hallucination}
\label{subsec:m-capjudge}

The CAPability eval already ships with a Qwen3-30B-A3B-Instruct-2507 judge whose per-example scores are persisted in the eval output. We use them, but for \textbf{recall}, not precision. The judge is given a single ground-truth object label (e.g.\ ``stain'', ``arch'') and asked whether the caption mentions this target. Score $= 1$ (yes), 0 (no), $-1$ (judge couldn't decide).

That's a coverage check. ``No'' means the model \emph{omitted} the target, not that the model hallucinated. The judge is not testing whether the caption made any false claims, only whether it made \emph{enough} claims to cover the target list.

Using this directly as a hallucination signal would be a mistake, and tempting because the scores are already computed. It would also push in the \emph{wrong direction}: verbose models naturally cover more targets and would appear less hallucinatory by this measure simply because they say more things. The same measurement is informative as \textbf{recall}, and that's how we use it: per-model fraction of capability targets correctly covered, category-balanced over the 9 capability dimensions, with the $-1$ rows excluded from the rate denominator. Together, DetailCaps LLM-judge precision and CAPability text-judge recall are the two halves of the brevity-doesn't-cost-anything claim.

\subsection{Per-example metrics for the verbosity dissociation}
\label{subsec:m-dissociation}

The \S\ref{sec:verbosity} dissociation runs on all 18 evaluations at 2B and 4B; seeds for the internal models are majority-pooled.

\textbf{Binary correctness.} For 17 of 18 evals we threshold the per-example score at 0.5 (binary outcomes pass through; the few continuous-in-$[0,1]$ outcomes, DocVQA, TextVQA, and PixMo Points, become binary at this cut), and drop the CAPability judge-undecided ($-1$) rows. For DetailCaps, where no rule-based per-example score is available, we use frontier LLM-judge precision and call a caption correct if precision $\geq 0.7$ ($\geq 70\%$ of the atomic factual claims supported by the human references, unverifiable claims excluded; \S\ref{subsec:m-judge}). \emph{Threshold knob}: at 0.5, 95--99\% of captions pass and models are indistinguishable; at 0.85, only 37--66\% pass; at 0.7 the pass rate is 70--93\%, the regime where models meaningfully differ, and the qualitative pattern is stable from 0.6 to 0.8.

\textbf{Mean accuracy and pairwise deltas} ($Q{-}D$, $M{-}D$, $Q{-}M$). Per (model, benchmark, scale), the mean of the per-example correctness label across examples. The three differences isolate distinct effects:
\begin{itemize}
  \item $\bm{Q - D}$: the deployment-relevant question, whether Qwen3.5 produces more correct answers per query than the \curated model at fixed scale.
  \item $\bm{M - D}$: the \textbf{verbosity-without-reasoning} effect (M emits ${\sim}60$ non-chain-of-thought tokens, D ${\sim}30$). If $M-D$ is negative everywhere, generic verbose tokens do not buy correctness.
  \item $\bm{Q - M}$: the \textbf{reasoning-content} effect controlling for verbosity (Q and M both emit non-trivial tokens; Q's are reasoning-structured).
\end{itemize}

\textbf{Regime classification.} Each (scale, capability) comparison is labeled by the pair $(Q-D, M-D)$ at a 2-pp threshold: \textbf{Reasoning helps} ($Q-D > 2$, $M-D$ within $\pm2$); \textbf{Reasoning helps, generic verbose hurts} ($Q-D > 2$, $M-D < -2$); \textbf{Concise wins} (both $< -2$); \textbf{Filler} (both within $\pm2$); and \textbf{Mixed / inconclusive} otherwise. Lowering the 2-pp cut sharpens the named-regime assignments but is noisier; raising it pushes more comparisons to ``Mixed.''
\clearpage
\renewcommand{\thesection}{A}
\section{Matched-scale captioning quality}
\label{app:captioning}

The blog's central positive claim is on the matched-scale controlled contrast: \textbf{Datology 2B vs Mammoth 2B}. Same Qwen3-1.7B backbone. Same VL training recipe. Same 4k-token context. Identical setup in every dimension except pretraining data composition. This isolates the \emph{curation} effect from scale, architecture, recipe, and context-length confounds.

On this pair the matched-scale captioning metrics line up (Table~\ref{tab:captioning-metrics}):

\begin{table}[h]
  \centering
  \small
  \begin{tabular}{l rrr}
    \toprule
    Captioning-specific metric (controlled pair, 2B) & Datology & Mammoth & $\Delta$ \\
    \midrule
    DetailCaps precision (LLM judge)               & 0.833 & 0.815 & $+0.018$ \\
    CAPability recall (category-balanced, 9 dims)  & 0.630 & 0.624 & $+0.006$ \\
    Mean DetailCaps caption length (chars)         & 993   & 2{,}449 & $-1{,}456$ ($-59\%$) \\
    Hallucinated mentions / caption                & 1.98  & 2.45  & $-0.47$ ($-19\%$ relative) \\
    \bottomrule
  \end{tabular}
  \caption{\textbf{Captioning-specific metrics on the controlled 2B pair.} Precision and recall favor Datology by a small but robust margin (precision $+0.018$ is ${\sim}4.5\times$ the cross-seed std; the per-seed bands don't overlap); the large magnitudes are in brevity (59\% shorter) and total hallucinated mentions (19\% fewer).}
  \label{tab:captioning-metrics}
\end{table}
At matched scale, curation produces a caption that is \emph{equal-or-better on the things we measure as quality} and \emph{substantially cheaper on the things we measure as cost}. That's the strict-Pareto-improvement claim, and it survives every methodology choice in \S\ref{app:methods}. The frontier 4B comparisons (Qwen3.5-4B, Qwen3-VL-4B, InternVL3.5-4B) are honest mixed pictures: verbose 4B models match or beat Datology 2B on the quality axes but at much higher per-token cost, and the main text treats them as such, not as wins.

The per-caption hallucinated-mention count decomposes into a length effect (longer outputs accumulate more hallucinations because there are more chances to err) and a per-token-rate effect (does each unit length carry more or fewer hallucinations?). For the curated--baseline contrast with $\Delta H = H_{\text{Datology}} - H_{\text{Mammoth}}$ and $H = n \cdot r$:
\begin{equation*}
  \Delta H = \underbrace{(n_{\text{Datology}} - n_{\text{Mammoth}}) \cdot r_{\text{Mammoth}}}_{\text{length effect}} + \underbrace{n_{\text{Datology}} \cdot (r_{\text{Datology}} - r_{\text{Mammoth}})}_{\text{per-token-rate effect}}.
\end{equation*}
On the controlled pair: length effect $= \mathbf{-1.46}$, per-token-rate effect $= \mathbf{+0.99}$, net $\Delta H = \mathbf{-0.47}$. At matched scale, the \curated model produces fewer total hallucinated mentions per caption \emph{by saying less, not by saying each thing more faithfully}; per character, its grounding is essentially Mammoth's. We are therefore \textbf{not} claiming that curation teaches the captioning model to hallucinate less per token. The claim is narrower: the accuracy improvement persists through length control on the 18-eval suite (the $+17.6$~pp pooled AME), and on the long-form captioning subset the hallucination win is explained by brevity.
\clearpage
\renewcommand{\thesection}{B}
\section{Output length: per-model distributions and accuracy splines}
\label{app:splines}

\begin{figure}[h]
  \centering
  \includegraphics[width=\linewidth]{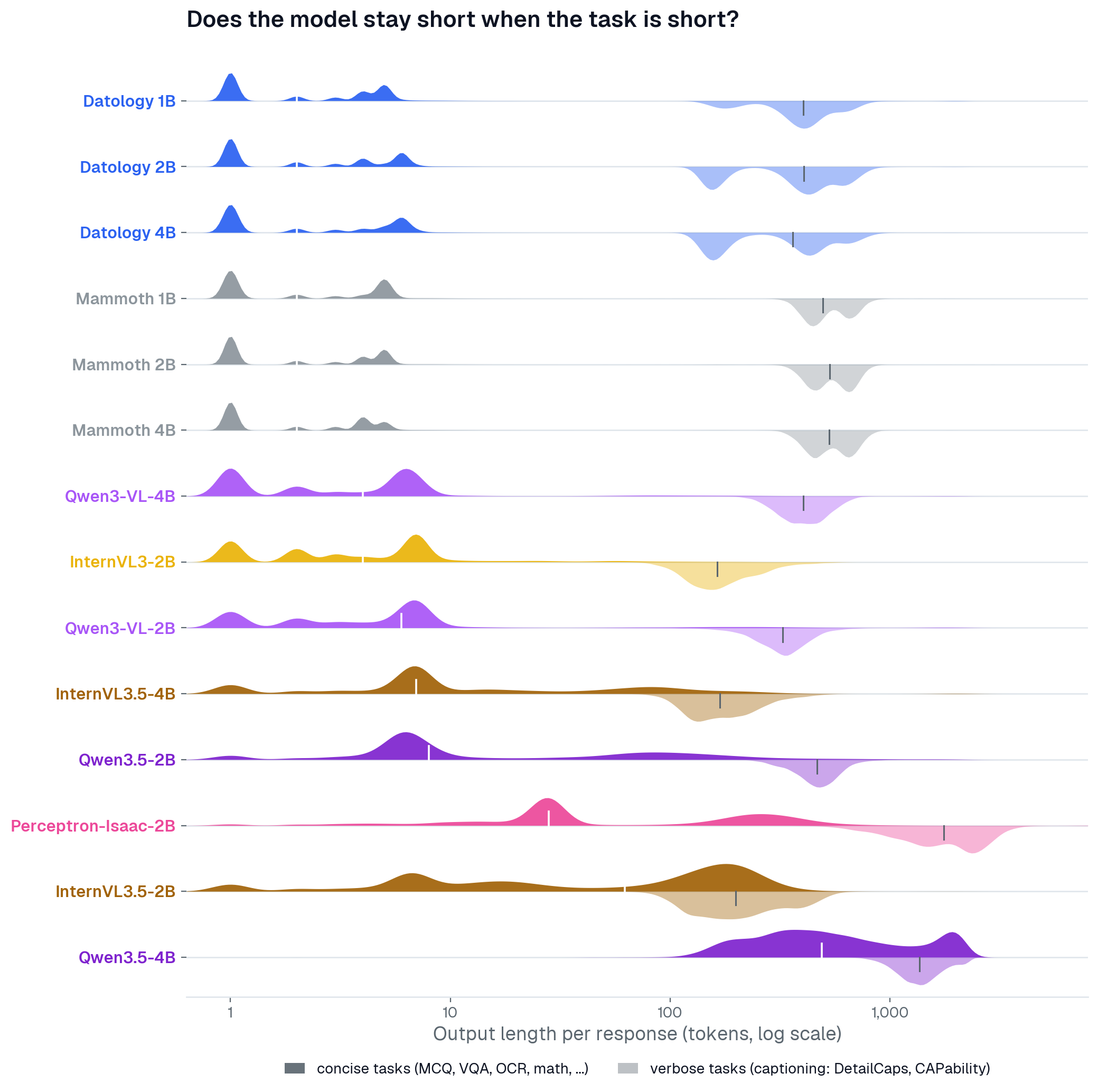}
  \caption{\textbf{Curated models stay concise when the task is concise; verbose models don't.} Per-response output-length distribution for every tested model, split by task type: concise tasks (MCQ, VQA, OCR, math, \ldots; top half of each violin) versus verbose captioning tasks (DetailCaps and CAPability; bottom half). Models are sorted by concise-task median; the white and grey ticks mark the concise and verbose medians. DatologyAI and Mammoth collapse to a few tokens on concise tasks and lengthen only for captioning, whereas the reasoning models stay long even when a single token would do: Qwen3.5-4B never drops below a few hundred tokens, which is why it has no concise operating point (\S\ref{sec:frontier}).}
  \label{fig:length-violin}
\end{figure}

\Cref{fig:length-violin} shows the per-response output-length distribution for every tested model, split by whether the task calls for a short answer or a long one. The curated models collapse to a few tokens on concise tasks and lengthen only for captioning, while the reasoning models stay long regardless --- the operating-length gap that the cost analysis (\S\ref{sec:cost}) turns into a Cost-of-Pass gap.

\begin{figure}[h]
  \centering
  \includegraphics[width=0.85\linewidth]{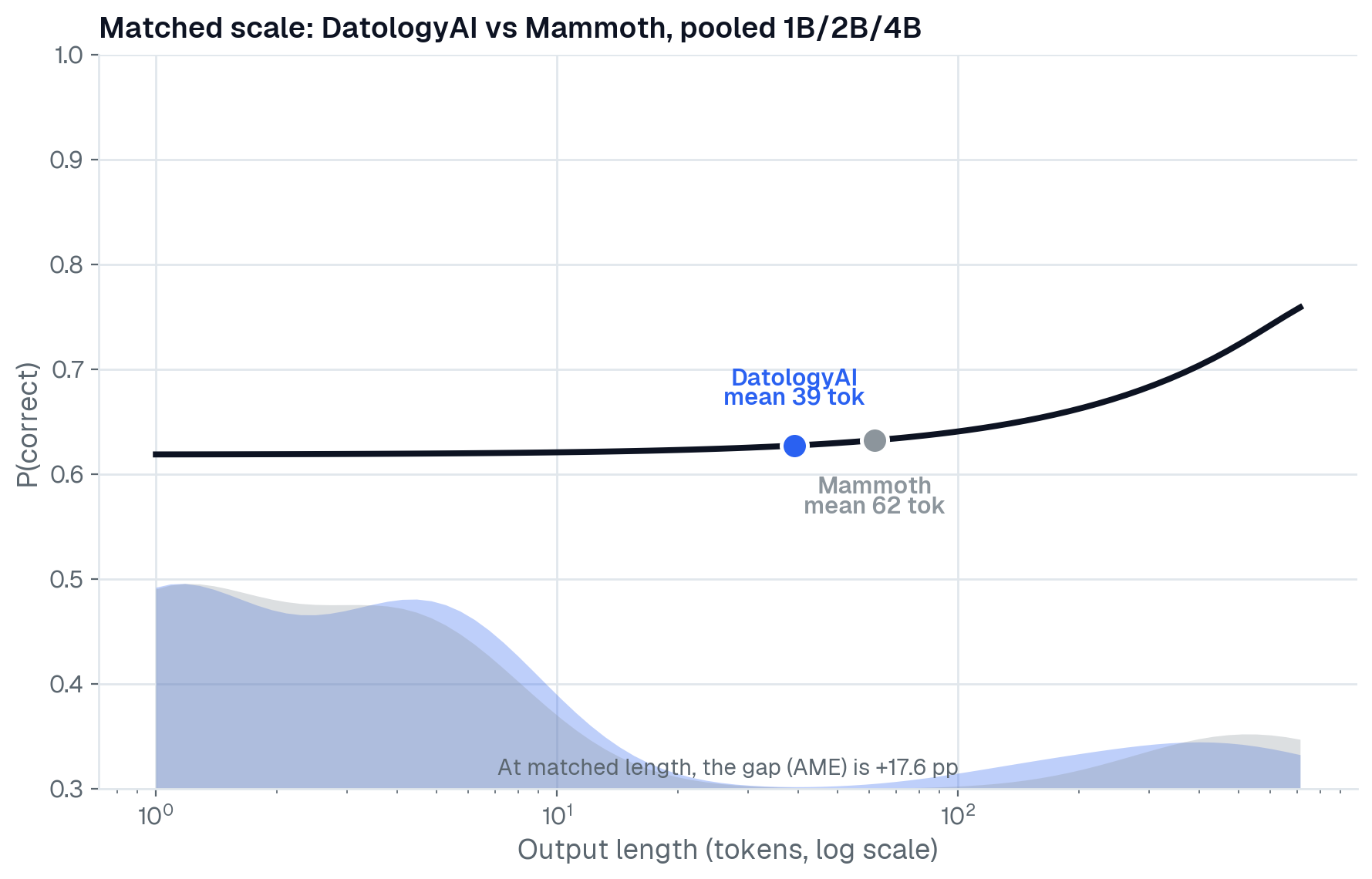}
  \caption{\textbf{At matched scale, the accuracy gap is model identity, not length.} Marginal $\Pr(\text{correct})$ against output length (black curve, controlling for benchmark); each model's mean output length (the 20-eval macro-mean, the same quantity the cost analysis uses) is a dot on the curve, and its full length distribution is the strip along the bottom. The black curve is pooled across both models: the GLM fits a single shared length spline plus an additive model term, so both follow the same curve shape and differ only by a model-level offset (the AME), not by how accuracy responds to length. DatologyAI averages ${\sim}39$ tokens and Mammoth ${\sim}62$ (the curated model is the shorter of the pair); both means sit on the flat part of the curve (the suite is dominated by short-answer benchmarks, with a long-form captioning tail), so the length difference contributes essentially nothing to the raw mean gap (raw $+17.1 \approx$ AME $+17.6$). The annotated AME is the model-identity coefficient, the curation effect with length held fixed, not a length artifact. The GLM is specified in \S\ref{subsec:m-quality}.}
  \label{fig:length-spline-controlled}
\end{figure}

\begin{figure}[h]
  \centering
  \includegraphics[width=\linewidth]{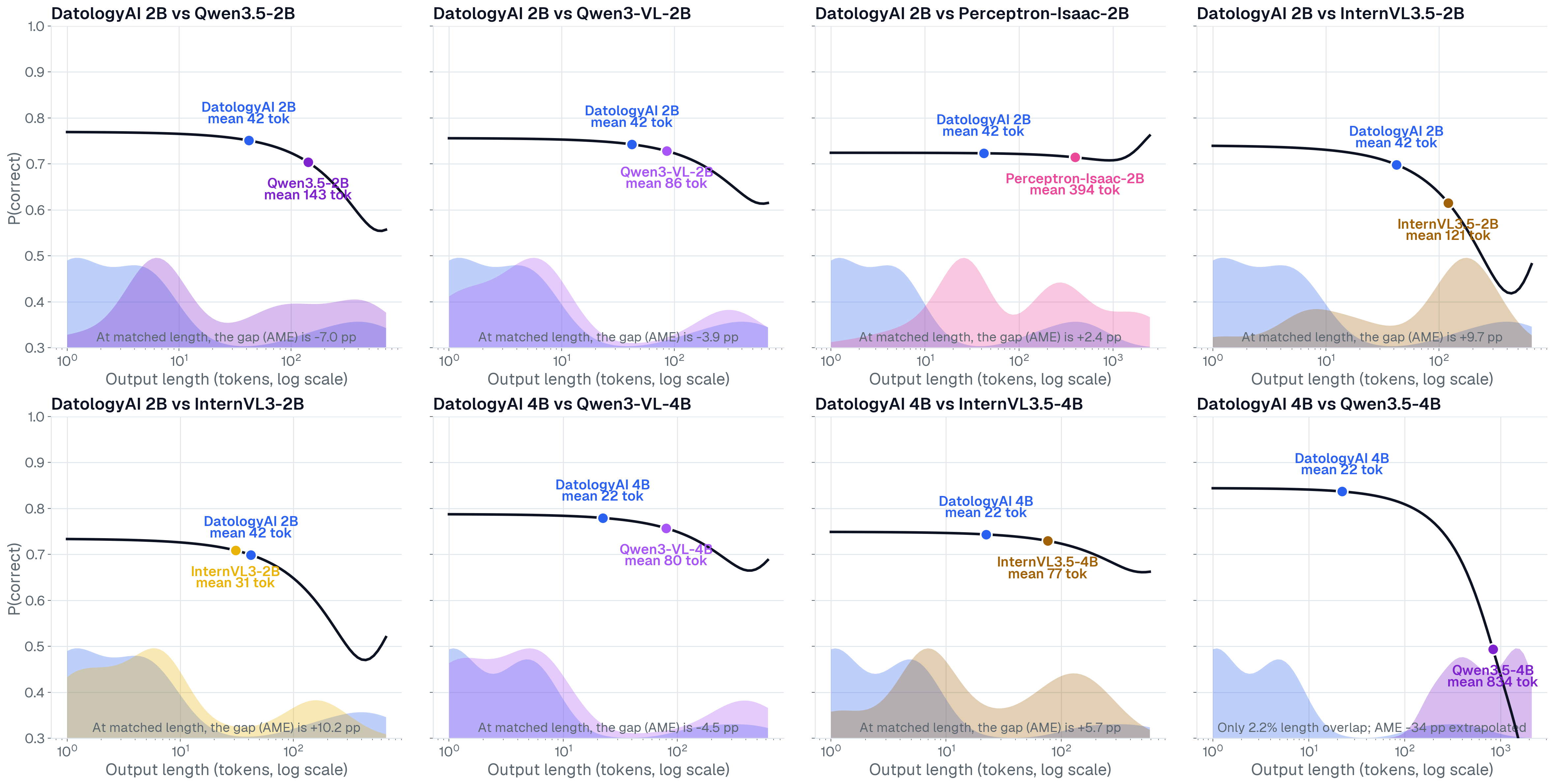}
  \caption{\textbf{Length-accuracy splines for all eight frontier comparators.} Each panel pairs the matched-scale \curated model (2B or 4B) with one external model: marginal $\Pr(\text{correct})$ against output length (black curve, controlling for benchmark), each model's mean output length (the 20-eval macro-mean) a dot on the curve, and its full length distribution the shaded strip along the bottom. In every panel the black curve is pooled across the two models shown (one shared length spline plus an additive model offset), so they differ by the AME, not by the shape of their length response. The annotated AME (the model-identity gap at matched length) ranges from $-7$~pp against Qwen3.5-2B to $+10$~pp against InternVL3-2B. The bottom-right pair, Datology 4B vs Qwen3.5-4B, is the one that fails the length-overlap test ($2.2\%$; \Cref{tab:positivity}): the two models barely share an operating range, so its AME is an extrapolation and the pair is excluded from the main-text comparison (\Cref{tab:frontier-cost}).}
  \label{fig:frontier-spline}
\end{figure}

\Cref{fig:length-spline-controlled,fig:frontier-spline} plot the fitted length--accuracy splines behind the matched-scale (\S\ref{sec:quality}) and frontier (\S\ref{sec:frontier}) length-controlled AMEs. The black curve in each is the length response pooled across the models shown (one shared spline plus an additive per-model offset), so the models differ by the AME, not by the shape of their length response; each model's mean output length is a dot on the curve, and its length distribution is the strip beneath.

\clearpage
\renewcommand{\thesection}{C}
\section{Refusal is not the mechanism: the length gap is response-conditional}
\label{app:refusal}

The observed Cost-of-Pass advantage rests on a lower mean output length. A reading consistent with those numbers but worse for the brevity story would be that the \curated model's lower mean arises because it refuses to answer more often. Refusals tend to be terse (e.g.\ ``I can't help with that'', ``I cannot see X from this image''), so a model that refuses an extra ten percentage points of the time would mechanically post a lower mean output length without changing its response-conditional behavior. Under that mechanism, \S\ref{sec:cost}'s cost advantage would describe a safety / behavior shift rather than inference efficiency in earnest.

This hypothesis is testable. The mean-output-length gap $\bar{n}_a - \bar{n}_b$ between the \curated model ($a$) and a comparator ($b$) decomposes additively into three terms:
\begin{equation*}
\begin{aligned}
  \bar{n}_a - \bar{n}_b ={} & \underbrace{(P_a(R) - P_b(R)) \cdot (E_b[\text{len}\,|\,R] - E_b[\text{len}\,|\,r])}_{\text{(i) refusal-rate shift}} \\
  & {}+ \underbrace{P_a(R) \cdot (E_a[\text{len}\,|\,R] - E_b[\text{len}\,|\,R])}_{\text{(ii) refusal length, given refusing}} \\
  & {}+ \underbrace{(1 - P_a(R)) \cdot (E_a[\text{len}\,|\,r] - E_b[\text{len}\,|\,r])}_{\text{(iii) response length, given responding}}
\end{aligned}
\end{equation*}
where $P(R)$ is the probability of refusing and $r$ denotes responding. Term (i) captures ``refuses more often''; (ii) captures ``refuses more tersely''; (iii) captures ``responds more concisely.'' Under the safety-shift reading, term (i) would dominate. Under the inference-efficiency reading, term (iii) would dominate.

Refusal classification combines a rule-based regex pass (matching twelve VLM-abstention patterns: ``I cannot see'', ``is not visible'', ``I'm sorry, but'', etc.) with a frontier LLM-judge cleanup that distinguishes genuine abstentions from substantive negative-existence answers that happen to contain refusal-like phrasing (e.g.\ ``there are no zebras in this image'', ``the text is not legible from this angle''). The cleanup matters more than expected: on the eval-suite sample the regex pass alone flagged ${\sim}13.5$k responses as refusals, while the LLM judge ultimately classified only ${\sim}5\%$ of those as genuine abstentions and the remaining ${\sim}95\%$ as substantive answers with hedge phrasing. Without the cleanup the decomposition would have wildly over-counted refusal rates.

\paragraph{Eval-suite decomposition: 98--100\% of the length gap is response-conditional.}
The primary sample is the full eval suite, the per-example responses from the same 18-eval frontier suite whose mean output lengths produce \S\ref{sec:cost}'s cost numbers. Each contrast operates on the slice of examples both models actually evaluated.

\begin{table}[h]
  \centering
  \footnotesize
  \resizebox{\linewidth}{!}{%
  \begin{tabular}{l r r r r r r r r r}
    \toprule
    Pair & $n$ paired & $P_a(R)$ & $P_b(R)$ & $\bar{n}_a$ & $\bar{n}_b$ & $\Delta$ len & (i) \% & (ii) \% & (iii) \% \\
    \midrule
    matched 1B      & 83{,}316 & 0.001\% & 0.011\% & 242 & 406   & $-164$   & $+0.02$ & $-0.01$ & \textbf{100.0} \\
    matched 2B      & 92{,}974 & 0.000\% & 0.011\% & 280 & 375   & $-94$    & $-0.04$ & $+0.00$ & \textbf{100.0} \\
    matched 4B      & 92{,}981 & 0.000\% & 0.193\% & 185 & 385   & $-199$   & $+0.11$ & $+0.00$ & \textbf{99.9} \\
    vs Qwen3.5-2B   & 93{,}086 & 0.000\% & 0.103\% & 280 & 580   & $-300$   & $+0.89$ & $+0.00$ & \textbf{99.1} \\
    vs Qwen3-VL-2B  & 92{,}998 & 0.000\% & 0.052\% & 280 & 328   & $-48$    & $+1.82$ & $+0.00$ & \textbf{98.2} \\
    vs Qwen3.5-4B   & 93{,}102 & 0.000\% & 0.424\% & 185 & 3{,}154 & $-2{,}969$ & $+0.65$ & $+0.00$ & \textbf{99.3} \\
    vs Qwen3-VL-4B  & 92{,}993 & 0.000\% & 0.042\% & 185 & 338   & $-153$   & $+0.36$ & $+0.00$ & \textbf{99.6} \\
    \bottomrule
  \end{tabular}%
  }
  \caption{\textbf{Three-term decomposition of the mean-output-length gap on the eval suite: response-conditional brevity (term iii) carries 98--100\% of every contrast.} Each row pairs the \curated model ($a$) against a comparator ($b$) on the examples both evaluated ($n$ paired). $P_a(R)$ and $P_b(R)$ are each model's refusal probability after LLM-judge cleanup of the regex pass; $\bar{n}_a$ and $\bar{n}_b$ are mean output lengths in characters; $\Delta$ len $= \bar{n}_a - \bar{n}_b$. Columns (i)--(iii) give the share of $\Delta$ len carried by the refusal-rate shift, refusal length given refusing, and response length given responding, respectively; shares may not sum to exactly 100 due to rounding.}
  \label{tab:refusal-eval}
\end{table}

The first thing \Cref{tab:refusal-eval} shows is mechanical: the \curated model essentially never refuses on standard evaluations. Across 81k probes at 4B, the LLM judge identifies zero genuine refusals from the \curated model; across 81k at 2B, also zero; at 1B, one. Comparator refusal rates are higher but still small (0.04\%--0.42\% across the seven contrasts).

With refusals that rare, the decomposition is one-sided. Term (iii), response-conditional brevity, carries between \textbf{98.2\% and 100\%} of the mean-output-length gap in every contrast. Term (i), the refusal-rate shift, contributes at most 1.8\% in either direction, and in three of seven contrasts it is \emph{negative}: the \curated model refuses slightly less often than the comparator, so the rate effect widens the length gap rather than closing it. Term (ii), conditional refusal length, is essentially zero in every contrast, because the \curated model's refusal probability is so low that the term's $P_a(R)$ multiplier extinguishes whatever per-refusal length difference exists.

In headline terms: Datology 4B is 199 characters shorter than Mammoth 4B per response on the eval suite, and 99.9\% of that gap is response-conditional. Against Qwen3.5-4B the gap is 2,969 characters, of which 99.3\% is response-conditional. On the actual eval traffic that produces \S\ref{sec:cost}'s cost numbers, the brevity advantage isn't bought via refusal; it is bought one shorter response at a time.

\paragraph{Refusal-enriched bundle: response-conditional brevity holds on the harder test.}
The eval suite contains few refusal-eliciting probes: the 18 evals are standard VLM tasks where any of the models studied refuses at well under 1\%. A skeptical reading would be that the absence of refusals here doesn't rule out a refusal-shift mechanism; perhaps the curated model would refuse more often on probes specifically designed to elicit refusals, and the standard evaluations are too soft a sample to surface that.

Testing that reading requires a sample enriched for refusals, which is what the 1,129-probe curated-comparison bundle provides. It was constructed during an April 2026 internal redteam exercise, with approximately 10\% of probes (the refusal-calibration and termination-budget groups) explicitly designed to elicit refusal-vs-respond decisions. Refusal rates on the bundle are an order of magnitude higher than on the eval suite: the \curated model refuses on 4.5\% of bundle probes, Mammoth on 3.2\%, Qwen3-VL-2B-Instruct on 15.2\%. On a sample deliberately built to expose a refusal-rate difference, the comparison is discriminating in a way the eval suite is not (\Cref{tab:refusal-bundle}).

\begin{table}[h]
  \centering
  \small
  \begin{tabular}{l r r}
    \toprule
    Decomposition term & Curated vs Mammoth ($n=1{,}093$) & Curated vs Qwen3-VL-2B ($n=513$) \\
    \midrule
    (i) Refusal-rate shift               & $+0.03$ chars ($-0.2\%$) & $+4.10$ chars ($-1.9\%$) \\
    (ii) Refusal length, given refusing  & $-5.29$ chars (25.6\%)   & $-12.20$ chars (5.6\%) \\
    (iii) Response length, given responding & \textbf{\boldmath$-15.43$ chars (74.6\%)} & \textbf{\boldmath$-201.20$ chars (92.4\%)} \\
    Total ($\bar{n}_a - \bar{n}_b$)      & \textbf{\boldmath$-20.68$ chars} & \textbf{\boldmath$-217.70$ chars} \\
    \bottomrule
  \end{tabular}
  \caption{\textbf{The same three-term decomposition on the 1,129-probe curated-comparison bundle, the refusal-enriched harder test: term (iii) still carries the majority of the gap.} Cell entries are each term's contribution to the mean-output-length gap in characters, with its share of the total gap in parentheses.}
  \label{tab:refusal-bundle}
\end{table}

Even on the enriched sample, term (i), the refusal-rate shift, contributes essentially zero ($-0.2\%$ vs Mammoth, $-1.9\%$ vs Qwen3-VL-2B). Term (ii), refusal length given refusing, carries 6--26\%, non-negligible but only because both models actually refuse with measurable frequency on calibration probes, raising the $P_a(R)$ multiplier from the eval-suite vanishing point. Term (iii), response-conditional brevity, carries the majority of the gap in both contrasts (75\% vs Mammoth, 92\% vs Qwen3-VL). The discriminating sample reproduces the eval-suite conclusion at lower numerical intensity.

Notably, the \curated--Mammoth refusal-rate gap is 1.3~pp (4.5\% vs 3.2\%), within the 95\% confidence interval of zero at $n=1{,}093$; and against Qwen3-VL the \curated model refuses \emph{less} often (4.5\% vs 15.2\%), so the rate effect actively opens the length gap rather than closing it. The length advantage in both contrasts is response-conditional.

A scope note: neither sample is representative of in-the-wild deployment traffic. The bundle is hand-curated for capability surfacing; the eval suite is a public-eval distribution. A model with a different refusal calibration might refuse much more often on real user queries than on either of these distributions. What the analysis here demonstrates is the narrower, sufficient claim: that \S\ref{sec:cost}'s \emph{length} advantage is not bought via refusal-rate shift on the populations the cost numbers were computed on. A future characterization on deployment traffic would address the population-rate question this section does not.

\begin{takeaway}
\textbf{The brevity advantage in \S\ref{sec:cost} is response-conditional, not refusal-driven.} Curation-induced brevity isn't ascribable to increased refusal; when the curated model does respond, the responses themselves are shorter.
\end{takeaway}
\clearpage
\renewcommand{\thesection}{D}
\section{Per-benchmark and per-capability result tables}
\label{app:tablea1}

This appendix collects \S\ref{sec:verbosity}'s supporting tables: the per-benchmark grid (\Cref{tab:a1}, below), the per-capability triple-dissociation table (\Cref{tab:dissociation-main}), and the per-example contingency rates (\Cref{tab:contingency}). In the grid, each row is one (scale, benchmark) cell; capability groups follow the \S\ref{subsec:m-pool} ordering and benchmarks within a group are sorted alphabetically by display name. Numbers are mean per-example correctness (per-example score thresholded at 0.5 for 17 evals; LLM-judge precision thresholded at 0.7 for DetailCaps; seeds majority-pooled for the internal models). Per-benchmark numbers are benchmark-pooled (each example weighted equally within a benchmark) and so differ slightly from the example-pooled capability numbers in \Cref{tab:contingency}.

\paragraph{Column key.}
\begin{itemize}
  \item \textbf{\boldmath$\cD$ / $\cM$ / $\cQ$} --- mean accuracy for \nDat, \nMam, \nQwen at the matched scale (\%, 0--100).
  \item \textbf{\boldmath$\cQ{-}\cD$ / $\cM{-}\cD$} --- percentage-point deltas vs \nDat (positive $=$ verbose model is more accurate on average).
  \item \textbf{\boldmath$\cQ_{\text{res}}$} (rescue) --- $P(\cQ \text{ correct} \mid \cD \text{ wrong})$; of examples where \nDat fails, the fraction \nQwen rescues (\%).
  \item \textbf{\boldmath$\cQ_{\text{mis}}$} (misled) --- $P(\cQ \text{ wrong} \mid \cD \text{ correct})$; of examples where \nDat succeeds, the fraction \nQwen loses (\%).
  \item \textbf{Regime} --- the \S\ref{subsec:m-dissociation} regime classification with a 2-pp threshold on $\cQ{-}\cD$ and $\cM{-}\cD$.
\end{itemize}

\begin{table}[h]
  \centering
  \resizebox{\textwidth}{!}{%
  \begin{tabular}{l l l r r r r r r r l}
    \toprule
    Scale & Benchmark & Capability & $\cD$ & $\cM$ & $\cQ$ & $\cQ{-}\cD$ & $\cM{-}\cD$ & $\cQ_{\text{res}}$ & $\cQ_{\text{mis}}$ & Regime \\
    \midrule
    2B & PixMo Points & Referring \& Grounding & 28.2 & 8.4  & 18.5 & $-9.6$  & $-19.8$ & 17.0 & 77.5 & Concise wins \\
    2B & RefCOCO      & Referring \& Grounding & 86.0 & 44.6 & 89.1 & $+3.1$  & $-41.3$ & 48.5 & 4.3  & Reasoning helps; generic verbose hurts \\
    2B & RefCOCO+     & Referring \& Grounding & 77.1 & 40.2 & 83.6 & $+6.5$  & $-37.0$ & 49.7 & 6.3  & Reasoning helps; generic verbose hurts \\
    2B & RefCOCOG     & Referring \& Grounding & 80.3 & 39.0 & 83.7 & $+3.4$  & $-41.4$ & 47.4 & 7.4  & Reasoning helps; generic verbose hurts \\
    2B & MMBench      & General VQA            & 70.2 & 70.0 & 78.0 & $+7.8$  & $-0.2$  & 44.7 & 7.9  & Reasoning helps \\
    2B & RealWorldQA  & General VQA            & 58.0 & 46.9 & 58.7 & $+0.7$  & $-11.1$ & 33.3 & 23.0 & Mixed / inconclusive \\
    2B & DocVQA       & OCR \& Document        & 88.2 & 84.2 & 91.1 & $+2.9$  & $-4.0$  & 67.2 & 5.7  & Reasoning helps; generic verbose hurts \\
    2B & OCRBench     & OCR \& Document        & 70.6 & 66.9 & 86.3 & $+15.7$ & $-3.7$  & 65.0 & 4.8  & Reasoning helps; generic verbose hurts \\
    2B & TextVQA      & OCR \& Document        & 74.6 & 69.9 & 69.5 & $-5.1$  & $-4.8$  & 32.8 & 18.0 & Concise wins \\
    2B & CAPability   & Captioning             & 67.2 & 66.9 & 80.3 & $+13.1$ & $-0.3$  & 47.5 & 3.6  & Reasoning helps \\
    2B & DetailCaps   & Captioning             & 90.1 & 85.2 & 79.9 & $-10.2$ & $-4.9$  & 59.2 & 17.8 & Concise wins \\
    2B & 3DSRBench    & Spatial \& 3D          & 45.6 & 31.1 & 47.0 & $+1.4$  & $-14.6$ & 29.0 & 31.6 & Mixed / inconclusive \\
    2B & CVBench2D    & Spatial \& 3D          & 65.5 & 46.5 & 62.9 & $-2.6$  & $-19.0$ & 30.6 & 20.1 & Concise wins \\
    2B & CVBench3D    & Spatial \& 3D          & 77.4 & 56.8 & 81.0 & $+3.6$  & $-20.7$ & 51.3 & 10.3 & Reasoning helps; generic verbose hurts \\
    2B & CountBench   & Counting               & 90.4 & 87.0 & 85.3 & $-5.1$  & $-3.5$  & 53.2 & 11.3 & Concise wins \\
    2B & AI2D         & Chart \& Diagram       & 74.0 & 71.1 & 77.4 & $+3.4$  & $-2.9$  & 52.5 & 13.9 & Reasoning helps; generic verbose hurts \\
    2B & ChartQA      & Chart \& Diagram       & 82.0 & 79.8 & 83.1 & $+1.1$  & $-2.2$  & 48.6 & 9.3  & Mixed / inconclusive \\
    2B & MathVista    & Math                   & 54.3 & 49.4 & 71.4 & $+17.1$ & $-4.9$  & 54.0 & 14.0 & Reasoning helps; generic verbose hurts \\
    \midrule
    4B & PixMo Points & Referring \& Grounding & 43.9 & 8.1  & 19.3 & $-24.6$ & $-35.8$ & 13.6 & 73.4 & Concise wins \\
    4B & RefCOCO      & Referring \& Grounding & 92.7 & 30.8 & 87.6 & $-5.1$  & $-62.0$ & 35.4 & 8.3  & Concise wins \\
    4B & RefCOCO+     & Referring \& Grounding & 88.2 & 26.8 & 83.6 & $-4.5$  & $-61.3$ & 37.9 & 10.2 & Concise wins \\
    4B & RefCOCOG     & Referring \& Grounding & 87.8 & 27.1 & 83.4 & $-4.4$  & $-60.6$ & 41.9 & 10.8 & Concise wins \\
    4B & MMBench      & General VQA            & 80.2 & 74.1 & 81.8 & $+1.5$  & $-6.1$  & 44.0 & 9.0  & Mixed / inconclusive \\
    4B & RealWorldQA  & General VQA            & 69.0 & 52.5 & 63.8 & $-5.2$  & $-16.5$ & 34.6 & 23.1 & Concise wins \\
    4B & DocVQA       & OCR \& Document        & 94.2 & 87.8 & 93.6 & $-0.6$  & $-6.4$  & 68.5 & 4.9  & Mixed / inconclusive \\
    4B & OCRBench     & OCR \& Document        & 78.5 & 69.0 & 89.3 & $+10.8$ & $-9.5$  & 68.4 & 5.0  & Reasoning helps; generic verbose hurts \\
    4B & TextVQA      & OCR \& Document        & 80.5 & 74.8 & 79.1 & $-1.4$  & $-5.7$  & 39.0 & 11.2 & Mixed / inconclusive \\
    4B & CAPability   & Captioning             & 73.8 & 68.1 & 80.5 & $+6.7$  & $-5.6$  & 44.6 & 6.8  & Reasoning helps; generic verbose hurts \\
    4B & DetailCaps   & Captioning             & 96.4 & 87.4 & 93.2 & $-3.2$  & $-9.1$  & 78.2 & 6.2  & Concise wins \\
    4B & 3DSRBench    & Spatial \& 3D          & 55.5 & 40.8 & 45.0 & $-10.5$ & $-14.7$ & 22.7 & 37.1 & Concise wins \\
    4B & CVBench2D    & Spatial \& 3D          & 75.8 & 53.3 & 72.4 & $-3.4$  & $-22.5$ & 24.4 & 12.3 & Concise wins \\
    4B & CVBench3D    & Spatial \& 3D          & 84.4 & 63.3 & 84.1 & $-0.3$  & $-21.1$ & 64.7 & 12.3 & Mixed / inconclusive \\
    4B & CountBench   & Counting               & 93.9 & 84.5 & 89.8 & $-4.1$  & $-9.4$  & 53.3 & 7.8  & Concise wins \\
    4B & AI2D         & Chart \& Diagram       & 83.6 & 75.4 & 76.5 & $-7.2$  & $-8.2$  & 46.6 & 17.7 & Concise wins \\
    4B & ChartQA      & Chart \& Diagram       & 87.2 & 82.2 & 82.3 & $-4.9$  & $-5.0$  & 42.2 & 11.8 & Concise wins \\
    4B & MathVista    & Math                   & 65.9 & 52.8 & 64.1 & $-1.8$  & $-13.1$ & 43.4 & 25.2 & Mixed / inconclusive \\
    \bottomrule
  \end{tabular}%
  }
  \caption{\textbf{\boldmath Per-benchmark mean accuracy ($\cD$, $\cM$, $\cQ$), pairwise deltas ($\cQ{-}\cD$, $\cM{-}\cD$), conditional rescue and misled rates ($\cQ_{\text{res}}$, $\cQ_{\text{mis}}$), and regime label, at both scales.}}
  \label{tab:a1}
\end{table}

\paragraph{Patterns visible at the benchmark level that the per-capability roll-up smooths over.}
\begin{itemize}
  \item \textbf{\boldmath PixMo Points is the largest single source of $\cQ_{\text{mis}}$} --- 73--78\% at both scales, by far the highest rates in the 36-cell grid. Chain-of-thought reasoning on a structured \texttt{<point x y>} output format is a near-pure cost: \cQ discards correct pointing answers it already knows. This sits behind the Referring \& Grounding group's overall trajectory.
  \item \textbf{The Reasoning-helps regime collapses with scale.} At 2B, eight benchmarks land in the ``Reasoning helps; generic verbose hurts'' or ``Reasoning helps'' categories; at 4B, only two do (CAPability and OCRBench). The shrinking-\cQ-wins story is visible per-benchmark, not just per-capability.
  \item \textbf{The captioning split is two benchmarks pulling in opposite directions} at both scales: CAPability (recall-style) puts \cQ ahead by $+13.1$ / $+6.7$~pp; DetailCaps (precision-style at threshold 0.7) puts \cD ahead by $+10.2$ / $+3.2$~pp. The Captioning group's averaged ``Mixed'' verdict is structurally this split, not a noisy null result.
  \item \textbf{\boldmath$\cQ_{\text{res}}$ is high almost everywhere} (above 25\% in 32 of 36 cells; above 40\% in 19 of 36), confirming the contingency-view observation from \S\ref{sec:verbosity} that \cQ and \cD have substantially different error sets across nearly every (scale, benchmark) pair --- the \emph{direction} of the mean-accuracy comparison turns on \cD's accuracy floor, not on how often \cQ and \cD reach different answers.
\end{itemize}

\paragraph{Pool-wide tokens per correct.}
Table~\ref{tab:tokens-per-correct} is the per-model tokens-per-correct ranking behind the OckScore figure (\Cref{fig:ockscore}) and the \S\ref{sec:cost} cost claim.

\begin{table}[h]
  \centering
  \small
  \begin{tabular}{l rr}
    \toprule
    Model & Tokens/correct & Macro accuracy \\
    \midrule
    \textbf{Datology 4B}          & \textbf{47.0}  & \textbf{0.691} \\
    \textbf{Datology 1B}          & \textbf{50.8}  & \textbf{0.638} \\
    InternVL3-2B                  & 50.9  & 0.589 \\
    \textbf{Datology 2B}          & \textbf{63.8}  & \textbf{0.660} \\
    Mammoth 2B                    & 103.2 & 0.541 \\
    Mammoth 4B                    & 108.6 & 0.549 \\
    Mammoth 1B                    & 113.3 & 0.505 \\
    InternVL3.5-4B                & 129.7 & 0.658 \\
    Perceptron-Isaac-2B           & 142.3 & 0.631 \\
    Qwen3-VL-2B-Instruct          & 150.7 & 0.674 \\
    Qwen3-VL-4B-Instruct          & 151.4 & 0.719 \\
    InternVL3.5-2B                & 229.6 & 0.560 \\
    Qwen3.5-2B                    & 243.6 & 0.695 \\
    \textbf{Qwen3.5-4B}           & \textbf{1{,}823} & 0.704 \\
    \bottomrule
  \end{tabular}
  \caption{\textbf{Tokens per correct answer} (mean tokens / macro accuracy) across the 1B--4B open-weight pool, per the OckBench convention. The curated models hold the low-token frontier (47--64 tokens per correct) at competitive accuracy; the most verbose comparator, Qwen3.5-4B, spends ${\sim}40\times$ more per correct answer.}
  \label{tab:tokens-per-correct}
\end{table}

\paragraph{Length non-overlap: Datology 4B vs Qwen3.5-4B.}
Table~\ref{tab:positivity} is the positivity check behind \S\ref{sec:frontier}: the two models almost never operate at comparable output lengths, so a length-controlled comparison of the pair is not identified.

\begin{table}[h]
  \centering
  \small
  \begin{tabular}{l rr}
    \toprule
    Per-model-token output length & Datology 4B & Qwen3.5-4B \\
    \midrule
    Median across all 18 grouped frontier evals                   & \textbf{3 tokens} & \textbf{626 tokens} \\
    5th--95th percentile range                                    & $[1, 376]$ & $[172, 2024]$ \\
    Fraction of Datology rows inside Qwen3.5-4B's $[p10, p90]$     & \textbf{6.7\%} & --- \\
    Fraction of Qwen3.5-4B rows inside Datology's $[p10, p90]$     & --- & \textbf{2.2\%} \\
    \bottomrule
  \end{tabular}
  \caption{\textbf{Datology 4B and Qwen3.5-4B have almost no length overlap.} Qwen3.5-4B is never observed at a concise operating point in our data; the positivity assumption a length-controlled comparison requires does not hold for this pair.}
  \label{tab:positivity}
\end{table}

\paragraph{Confidence intervals for the frontier AMEs.}
95\% confidence intervals for the per-comparator length-controlled AMEs in \Cref{tab:frontier-cost}.

\begin{table}[h]
  \centering
  \small
  \begin{tabular}{l r l}
    \toprule
    Comparator & AME (pp) & 95\% CI \\
    \midrule
    vs Qwen3.5-2B            & $-6.99$  & $[-7.49, -6.50]$ \\
    vs Qwen3-VL-2B-Instruct  & $-3.93$  & $[-4.02, -3.85]$ \\
    vs Perceptron-Isaac-2B   & $+2.39$  & $[+1.94, +2.84]$ \\
    vs InternVL3.5-2B        & $+9.68$  & $[+8.93, +10.44]$ \\
    vs InternVL3-2B          & $+10.23$ & $[+10.06, +10.39]$ \\
    vs Qwen3-VL-4B-Instruct  & $-4.53$  & $[-4.86, -4.21]$ \\
    vs InternVL3.5-4B        & $+5.69$  & $[+5.32, +6.05]$ \\
    \bottomrule
  \end{tabular}
  \caption{\textbf{95\% confidence intervals for the frontier-pool length-controlled AMEs} (companion to \Cref{tab:frontier-cost}).}
  \label{tab:frontier-ci}
\end{table}

\paragraph{Per-capability triple dissociation (full table).}
Table~\ref{tab:dissociation-main} is the per-capability companion to \Cref{fig:verbosity-bars}: mean accuracy and the dissociation deltas at both scales, with the regime label per cell.

\begin{table}[h]
  \centering
  \footnotesize
  \setlength{\tabcolsep}{4pt}
  \begin{tabular}{l l r r r r r p{3.0cm}}
    \toprule
    Scale & Capability & \cD & \cM & \cQ & $\cQ{-}\cD$ & $\cM{-}\cD$ & Regime \\
    \midrule
    \textbf{2B} & Math (MathVista)        & 0.543 & 0.494 & 0.714 & \textbf{+17.1} & $-4.9$ & Reasoning helps; generic verbose hurts \\
    \textbf{2B} & OCR \& Document         & 0.778 & 0.737 & 0.823 & $+4.5$ & $-4.1$ & Reasoning helps; generic verbose hurts \\
    \textbf{2B} & General VQA             & 0.641 & 0.585 & 0.683 & $+4.2$ & $-5.7$ & Reasoning helps; generic verbose hurts \\
    \textbf{2B} & Chart \& Diagram        & 0.780 & 0.755 & 0.803 & $+2.2$ & $-2.6$ & Reasoning helps; generic verbose hurts \\
    \textbf{2B} & Captioning              & 0.786 & 0.760 & 0.801 & $+1.5$ & $-2.6$ & Mixed / inconclusive \\
    \textbf{2B} & Referring \& Grounding  & 0.679 & 0.330 & 0.687 & $+0.8$ & \textbf{\boldmath$-34.9$} & Mixed (\cQ$\approx$\cD; \cM collapses) \\
    \textbf{2B} & Spatial \& 3D           & 0.629 & 0.448 & 0.636 & $+0.8$ & $-18.1$ & Mixed (\cQ$\approx$\cD; \cM collapses) \\
    \textbf{2B} & Counting                & 0.904 & 0.870 & 0.853 & \textbf{\boldmath$-5.1$} & $-3.5$ & \textbf{Concise wins} \\
    \midrule
    \textbf{4B} & OCR \& Document         & 0.844 & 0.772 & 0.873 & $+2.9$ & $-7.2$ & Reasoning helps; generic verbose hurts \\
    \textbf{4B} & Captioning              & 0.851 & 0.777 & 0.868 & $+1.8$ & $-7.3$ & Mixed / inconclusive \\
    \textbf{4B} & Math (MathVista)        & 0.659 & 0.528 & 0.641 & $-1.8$ & $-13.1$ & Mixed / inconclusive \\
    \textbf{4B} & General VQA             & 0.746 & 0.633 & 0.728 & $-1.9$ & $-11.3$ & Mixed / inconclusive \\
    \textbf{4B} & Counting                & 0.939 & 0.845 & 0.898 & \textbf{\boldmath$-4.1$} & $-9.4$ & \textbf{Concise wins} \\
    \textbf{4B} & Spatial \& 3D           & 0.719 & 0.525 & 0.672 & \textbf{\boldmath$-4.7$} & $-19.4$ & \textbf{Concise wins} \\
    \textbf{4B} & Chart \& Diagram        & 0.854 & 0.788 & 0.794 & \textbf{\boldmath$-6.0$} & $-6.6$ & \textbf{Concise wins} \\
    \textbf{4B} & Referring \& Grounding  & 0.781 & 0.232 & 0.685 & \textbf{\boldmath$-9.6$} & \textbf{\boldmath$-54.9$} & \textbf{Concise wins} \\
    \bottomrule
  \end{tabular}
  \caption{\textbf{Per-capability triple dissociation.} $\cD =$ \nDat, $\cM =$ \nMam, $\cQ =$ \nQwen at the matched scale. Numbers are mean per-example correctness (per-example score thresholded at 0.5 for 17 evals; LLM-judge precision thresholded at 0.7 for DetailCaps; seeds majority-pooled). Within-capability values are benchmark-averaged (each eval weighted equally). Regime labels follow the \S\ref{subsec:m-dissociation} rules with a 2-pp threshold.}
  \label{tab:dissociation-main}
\end{table}

\paragraph{Per-example contingency: rescue and misled rates.}
Per (benchmark, scale, example) each model's binary correctness sits in one cell of $\{\cD, \cM, \cQ\} \in \{0,1\}^3$; the marginal is mean accuracy, but the conditional rates expose structure the marginal hides. For the (\cD, \cQ) contingency we report the \textbf{rescue rate} $P(\cQ \text{ correct} \mid \cD \text{ wrong})$ (of \cD's misses, the fraction \cQ gets right) and the \textbf{misled rate} $P(\cQ \text{ wrong} \mid \cD \text{ correct})$ (of \cD's hits, the fraction \cQ loses). The mean-accuracy gap follows from them by the identity
\begin{equation*}
  \cQ_{\text{acc}} - \cD_{\text{acc}} = \underbrace{\cQ_{\text{rescue}} \times (1 - \cD_{\text{acc}})}_{\text{lift on D's misses}} - \underbrace{\cQ_{\text{misled}} \times \cD_{\text{acc}}}_{\text{loss on D's hits}}.
\end{equation*}

\begin{table}[h]
  \centering
  \small
  \begin{tabular}{l r r r r r r}
    \toprule
    Capability & $\cD_{\text{acc}}$ & $\cQ_{\text{acc}}$ & $\cQ_{\text{rescue}}$ & $\cQ_{\text{misled}}$ & $\cQ_{\text{rescue}}{-}\cQ_{\text{misled}}$ & $\cQ_{\text{acc}}{-}\cD_{\text{acc}}$ \\
    \midrule
    Counting               & 0.939 & 0.898 & \textbf{53.3\%} & 7.8\%  & $+45.5$~pp & \textbf{\boldmath$-4.1$~pp} \\
    OCR \& Document        & 0.868 & 0.868 & \textbf{49.3\%} & 7.5\%  & $+41.8$~pp & 0.0~pp \\
    Captioning             & 0.829 & 0.856 & \textbf{47.5\%} & 6.5\%  & $+41.0$~pp & $+2.7$~pp \\
    Chart \& Diagram       & 0.852 & 0.791 & 44.9\% & 15.0\% & $+29.9$~pp & \textbf{\boldmath$-6.1$~pp} \\
    Math                   & 0.659 & 0.641 & 43.4\% & 25.2\% & $+18.2$~pp & $-1.8$~pp \\
    General VQA            & 0.785 & 0.791 & 42.0\% & 10.8\% & $+31.2$~pp & $+0.5$~pp \\
    Referring \& Grounding & 0.884 & 0.834 & 37.2\% & 10.5\% & $+26.7$~pp & \textbf{\boldmath$-5.0$~pp} \\
    Spatial \& 3D          & 0.637 & 0.561 & 25.7\% & 26.6\% & $-0.9$~pp  & \textbf{\boldmath$-7.6$~pp} \\
    \bottomrule
  \end{tabular}
  \caption{\textbf{Per-capability rescue and misled rates at 4B.} Accuracies and conditional rates are example-pooled within capability (each example weighted equally), so they differ slightly from \Cref{tab:dissociation-main}'s benchmark-pooled accuracies; the qualitative ordering is unchanged.}
  \label{tab:contingency}
\end{table}

On 7 of 8 capability groups Qwen3.5-4B's rescue rate exceeds its misled rate, yet on five of those mean accuracy still says $\cQ \leq \cD$: when $\cD_{\text{acc}}$ is high the ``\cD-correct'' base is large and the ``\cD-wrong'' base is small, so even a small misled rate outweighs a large rescue rate. Counting at 4B is the worked case: $\cD_{\text{acc}} = 0.939$, $\cQ_{\text{rescue}} = 53.3\%$, $\cQ_{\text{misled}} = 7.8\%$, giving a lift of $53.3\% \times 0.061 = +3.3$~pp and a loss of $7.8\% \times 0.939 = -7.3$~pp, net $-4.1$~pp, exactly the mean-accuracy gap. The same rates with $\cD_{\text{acc}} = 0.65$ would instead yield a $+13.6$~pp \emph{gain}; only the population mix changed.

\end{document}